\documentclass[10pt,twocolumn,letterpaper]{article}

\usepackage{iccv}
\usepackage{times}
\usepackage{epsfig}
\usepackage{graphicx}
\usepackage{amsmath}
\usepackage{amssymb}

\usepackage{booktabs}
\usepackage{tablefootnote}
\usepackage[flushleft]{threeparttable}

\usepackage{lipsum}
\usepackage{bbding}
\usepackage{url}

\usepackage{color}
\usepackage{xcolor}
\usepackage{enumitem}

\usepackage[accsupp]{axessibility}

\definecolor{mypink3}{cmyk}{0, 0.7808, 0.4429, 0.1412}
\definecolor{myblue2}{cmyk}{0, 0., 0.4412, 0.4808}

\definecolor{myblue}{cmyk}{0, 0.7808, 0., 0.1412}

\usepackage[pagebackref=true,breaklinks=true,letterpaper=true,colorlinks,bookmarks=false]{hyperref}

\iccvfinalcopy 

\def\iccvPaperID{2535} 

\ificcvfinal\fi

\begin{document}

\title{Mask-Attention-Free Transformer for 3D Instance Segmentation}

\author{Xin Lai$^{1}$\hspace{1.0cm}Yuhui Yuan$^{3}$\hspace{1.0cm}Ruihang Chu$^{1}$\hspace{1.0cm}Yukang Chen$^{1}$\hspace{1.0cm}Han Hu$^{3}$\hspace{1.0cm}Jiaya Jia$^{1,2}$\thanks{Corresponding Author}\\
$^{1}$The Chinese University of Hong Kong~~~
$^{2}$SmartMore~~~
$^{3}$Microsoft Research Asia~~~
}

\maketitle
\ificcvfinal\thispagestyle{empty}\fi

\begin{abstract}
Recently, transformer-based methods have dominated 3D instance segmentation, where mask attention is commonly involved. Specifically, object queries are guided by the initial instance masks in the first cross-attention, and then iteratively refine themselves in a similar manner. However, we observe that the mask-attention pipeline usually leads to slow convergence due to low-recall initial instance masks. Therefore, we abandon the mask attention design and resort to an auxiliary center regression task instead. Through center regression, we effectively overcome the low-recall issue and perform cross-attention by imposing positional prior. To reach this goal, we develop a series of position-aware designs. First, we learn a spatial distribution of 3D locations as the initial position queries. They spread over the 3D space densely, and thus can easily capture the objects in a scene with a high recall. Moreover, we present relative position encoding for the cross-attention and iterative refinement for more accurate position queries. Experiments show that our approach converges $4\times$ faster than existing work, sets a new state of the art on ScanNetv2 3D instance segmentation benchmark, and also demonstrates superior performance across various datasets. Code and models are available at \url{https://github.com/dvlab-research/Mask-Attention-Free-Transformer}.
\end{abstract}

\section{Introduction}
Nowadays 3D point clouds can be conveniently collected. They have benefited various applications, such as autonomous driving, robotics, and augmented reality. As a fundamental task, 3D instance segmentation also poses great challenges simultaneously, such as geometric occlusion and semantic ambiguity.

Many works have been proposed to solve the 3D instance segmentation task. Grouping-based methods~\cite{jiang2020pointgroup,vu2022softgroup,chen2021hierarchical,zhong2022maskgroup} rely on heuristic clustering algorithms such as DBSCAN or Breadth-First Search (BFS) to generate instance proposals. They thus require sophisticated hyper-parameters tuning and are prone to wrongly segment instances that are close to each other. Recently, transformer-based methods~\cite{Schult23mask3d,sun2022superpoint} develop a fully end-to-end pipeline. With transformer decoder layers, a fixed number of object queries attend to global features iteratively and directly output instance predictions. It requires no post-processing for duplicate removal such as NMS, since it adopts one-to-one bipartite matching during training. Moreover, it employs mask attention, which uses the instance masks predicted in the last layer to guide the cross-attention. 

\label{sec:intro}

\begin{figure}
\begin{center}
\includegraphics[width=0.95\linewidth]{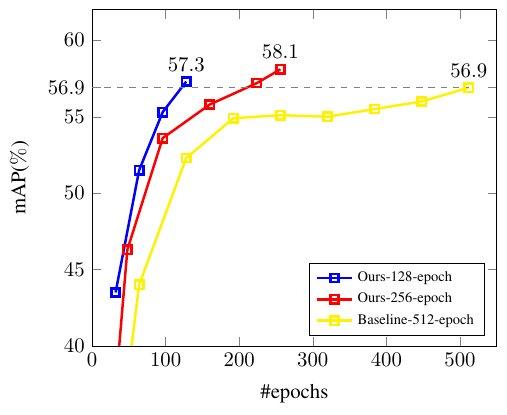}
\end{center}
\vspace{-0.5cm}
\caption{Validation curve of the baseline and ours on ScanNetv2 \textit{val} set. With only 128-epoch training, ours outperforms the baseline trained with 512 epochs.}
\label{fig:comp_curve}
\vspace{-0.3cm}
\end{figure}

\begin{figure}
\begin{center}
\includegraphics[width=1.0\linewidth]{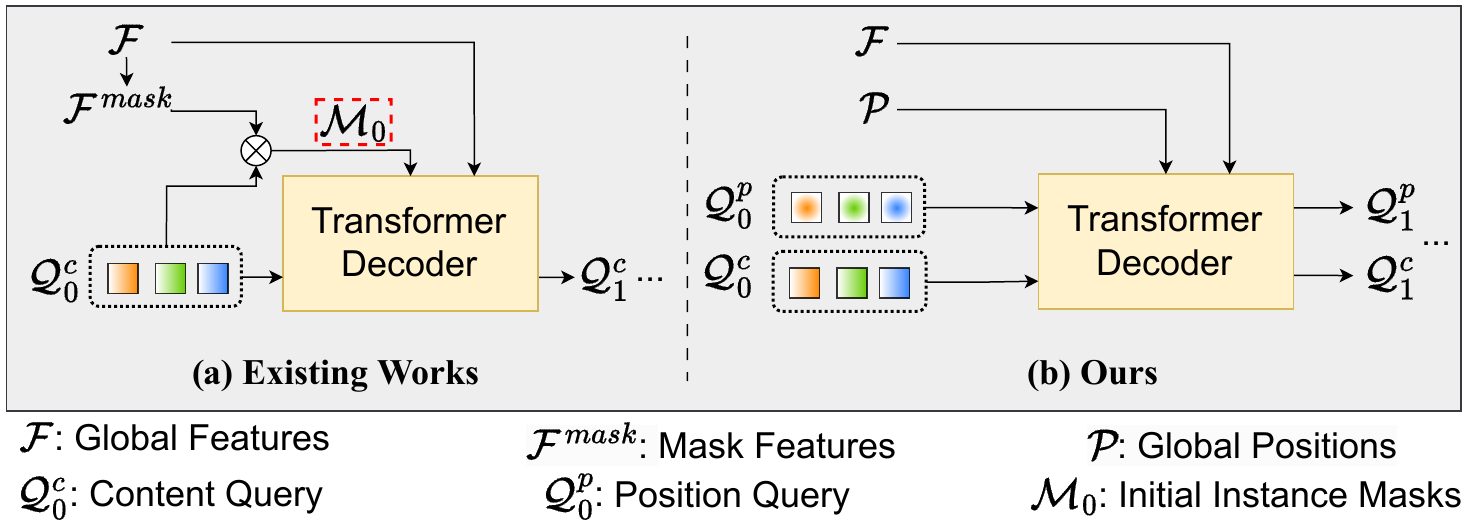}
\end{center}
\vspace{-0.3cm}
\caption{The framework of (a) existing works (based on mask attention) and (b) ours. Existing works have the issue of low-recall initial instance masks (\ie, $\mathcal{M}_0$). Our approach resorts to an auxiliary center regression task to circumvent this issue.}
\label{fig:comp_framework}
\vspace{-0.1cm}
\end{figure}

\begin{figure}
\begin{center}
\includegraphics[width=0.95\linewidth]{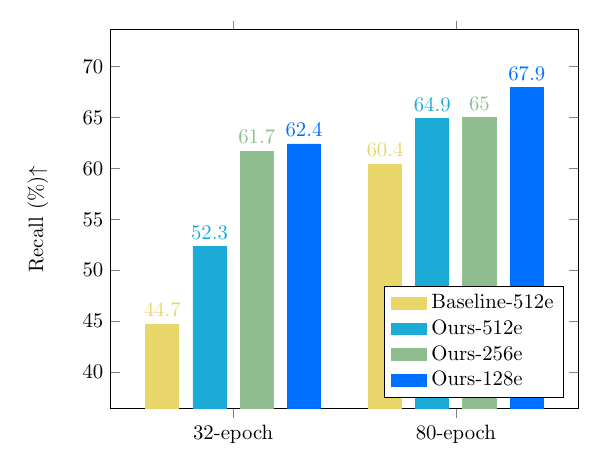}
\end{center}
\vspace{-0.7cm}
\caption{The recall of initial instance masks at the 32-th and 80-th training epoch. We compare the baseline trained with 512 epochs and ours trained with 128, 256, and 512 epochs. Since our approach does not produce initial instance masks before entering decoder layers, we make statistics on the instance masks output by the first decoder layer for both the baseline and ours.}
\label{fig:comp_recall}
\vspace{-0.2cm}
\end{figure}

However, we point out that current transformer-based methods suffer from the issue of slow convergence. As shown in Fig.~\ref{fig:comp_curve}, the baseline model manifests slow convergence and lags behind our method by a large margin, particularly in the early stages of training. We dive further and find that the issue is potentially caused by the low recall of the \textit{initial instance masks}. Specifically, as shown in Fig.~\ref{fig:comp_framework} (a), the initial instance masks are produced by the similarity map between the initial object queries and the point-wise mask features. Since the initial object queries are unstable in early training, we notice that the recall of initial instance masks is substantially lower than ours in Fig.~\ref{fig:comp_recall}, especially at the beginning of training (\ie, the 32-th epoch). The low-quality initial instance masks increase the training difficulty, thereby slowing down convergence.

Given the low recall of the initial instance masks, we abandon the mask attention design and instead construct an auxiliary center regression task to guide cross-attention, as depicted in Fig.~\ref{fig:comp_framework} (b). To enable center regression, we develop a series of position-aware designs. Firstly, we maintain a set of learnable position queries, each of which denotes the position of its corresponding content query. They are densely distributed over the 3D space, and we require each query to attend to its local region. As a result, the queries can easily capture the objects in a scene with a higher recall, which is crucial in reducing training difficulty and accelerating convergence. 

In addition, we design the contextual relative position encoding for cross-attention. Compared to the mask attention used in previous works, our solution is more flexible since the attention weights are adjusted by relative positions instead of hard masking. Furthermore, we iteratively update the position queries to achieve more accurate representation. Finally, we introduce the center distances between predictions and ground truths in both matching and loss. 

In total, our contribution is three-fold.
\begin{itemize}
    \item We observe that existing transformer-based methods suffer from the low recall of initial instance masks, which causes training difficulty and slow convergence.
    \item Instead of relying on mask attention, we construct an auxiliary center regression task to overcome the low-recall issue and design a series of position-aware components accordingly. Our approach manifests faster convergence and demonstrates higher performance.
    \item Experiments show our approach achieves a new state-of-the-art result and demonstrates superior performance on various datasets including ScanNetv2, ScanNet200, and S3DIS.
\end{itemize}

\begin{figure*}[ht]
\begin{center}
\includegraphics[width=1.0\linewidth]{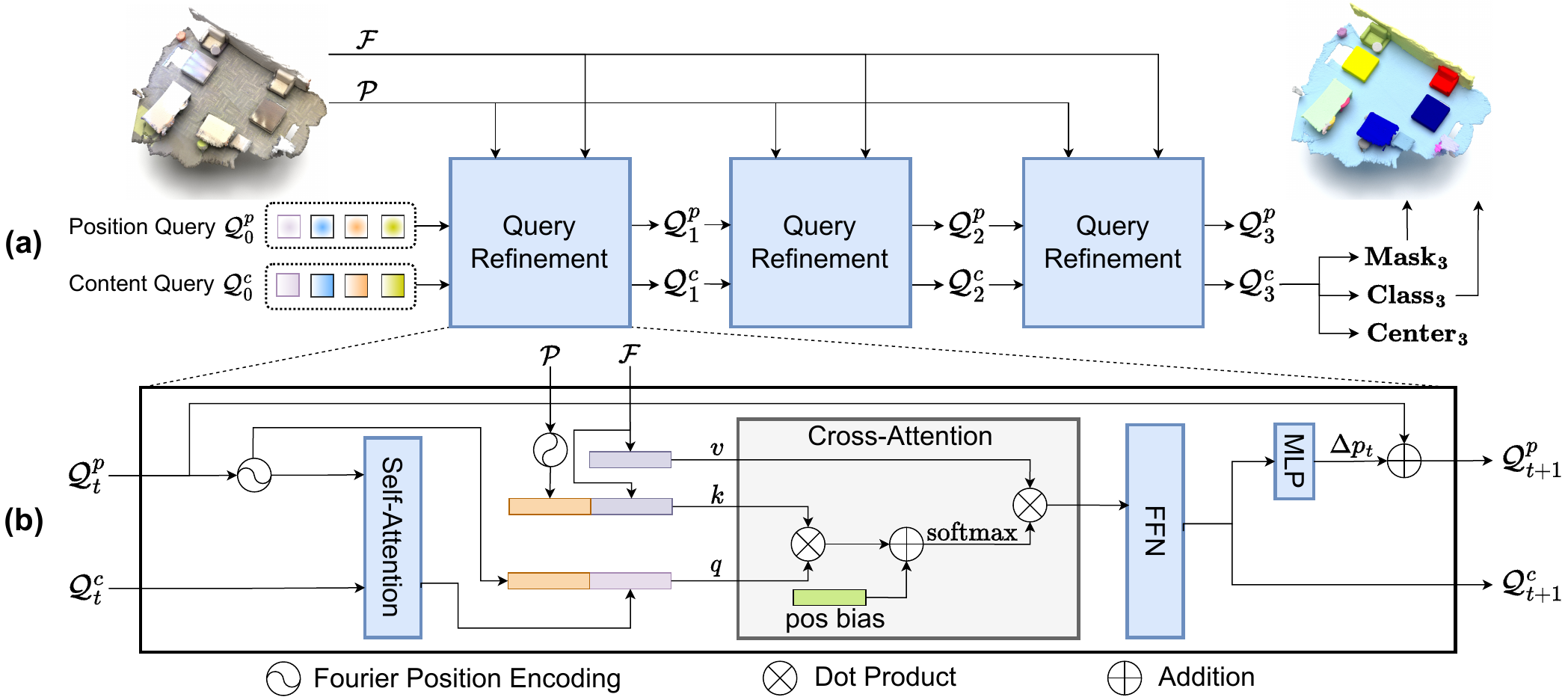}
\end{center}
\vspace{-0.4cm}
\caption{The overview of our framework. Besides content queries $\mathcal{Q}^c_{0}$, we also maintain a set of learnable position queries $\mathcal{Q}^p_{0}$. The content queries $\mathcal{Q}^c$ aggregate features from the global features $\mathcal{F}$. The position queries $\mathcal{Q}^p$ are designed to guide the cross-attention. The attention weights are adjusted based on the relative positions between the position queries $\mathcal{Q}^p$ and the global positions $\mathcal{P}$. Both content and position queries are iteratively refined in each layer. Only 3 decoder layers are shown (we use 6 layers in our experiments). }
\label{fig:overview}
\vspace{-0.1cm}
\end{figure*}

\vspace{-0.3cm}
\section{Related Work}

\subsection{3D Instance Segmentation} 3D instance segmentation is a fundamental task for 3D recognition~\cite{qi2017pointnet,qi2017pointnet++,lai2022stratified,chen2022focal,li2022voxel,liu2022spatial,li2022unifying,liu2023prior,lai2023spherical,jiang2021guided,liu2023mars3d,chu2023command}. The solutions can be categorized into detection-based, grouping-based and the new emerging transformer-based paradigms.
Detection-based approaches~\cite{hou20193d,YangWCHWMT19,yi2019gspn} first detect the bounding boxes and then segment the fine-grained instance mask. On the other hand, grouping-based approaches~\cite{wang2018sgpn,wang2019associatively,lahoud20193d,engelmann20203d,han2020occuseg,jiang2020pointgroup,jiang2020end,he2021dyco3d,zhang2021point,liang2021instance,chen2021hierarchical,vu2022softgroup,chu2022twist,chu2021icm} employ clustering algorithms to group the points into a set of instance clusters. Before clustering, they either move 3D points towards the associated object center to form a more compact distribution~\cite{lahoud20193d,han2020occuseg,jiang2020pointgroup,jiang2020end,zhang2021point,liang2021instance,chen2021hierarchical,vu2022softgroup}, or transform points to a high-dimension feature space~\cite{wang2018sgpn,wang2019associatively}. Further, a series of works leverage semantic priors to avoid the noisy points from other categories~\cite{lahoud20193d,jiang2020pointgroup,liang2021instance,chen2021hierarchical,vu2022softgroup} or use advanced grouping strategies~\cite{vu2022softgroup,chen2021hierarchical,liang2021instance}. With these designs, the grouping-based paradigm has achieved leading performance across various evaluation benchmarks\cite{armeni_cvpr16,dai2017scannet} for a long time.
Recently, transformer-based paradigm~\cite{Schult23mask3d,sun2022superpoint} becomes another option and swiftly sets a new state-of-the-art. Compared with previous methods, it presents an elegant pipeline, and can directly output instance predictions. It relies on the transformer decoder and mask attention to aggregate information from global features. In this work, we construct an auxiliary center regression task to assist in cross-attention. Although the existing grouping-based methods also predict center offsets, we explain that ours is not for instance proposals, but to overcome the low-recall issue and provide positional priors for cross-attention. 

\subsection{Vision Transformer}Transformer has become a fundamental model in the vision area, thanks to its flexibility and power to model various scenarios using attention mechanisms \cite{vaswani2017attention}. Recently, many works~\cite{dosovitskiy2020vit,pmlr-v139-touvron21a,touvron2021cait,wang2021pyramid,wang2021pvtv2,chu2021Twins,dong2021cswin,vip,yang2021focal} rely on the self-attention in transformers to develop vision fundamental models. Besides, DETR~\cite{detr} proposes a fully end-to-end pipeline for object detection. It utilizes transformer decoders to dynamically aggregate features from images, and uses one-to-one bipartite matching for ground-truth assignment, yielding an elegant pipeline. To solve the notorious slow convergence of DETR, approaches~\cite{zhu2020deformable,wang2022anchor,meng2021conditional,liudab,zhang2022accelerating,yu2022k} propose deformable attention, impose strong prior or decrease searching space in cross-attention to accelerate convergence. Further, methods of~\cite{li2022dn,liu2021group,jia2022detrs,zhang2022dino,li2022mask} present several ways to stabilize matching and training. Moreover, masked attention~\cite{cheng2021per,cheng2022masked} are proposed to impose semantic priors to accelerate training for segmentation tasks. Recently, there are works~\cite{lai2022stratified,zhao2021point,wu2022pointconvformer,wu2022point,park2022fast,Schult23mask3d,sun2022superpoint} that develop transformer models tailored for 3D point clouds. Following this line of research, we observe the low recall of initial instance masks, and present solutions to circumvent the use of mask attention.

\section{Method}

We first review previous methods and present the overview of our method in Sec.~\ref{sec:overview}. Then, we elaborate on the details of our position-aware designs in Sec.~\ref{sec:query_refinement_module}. 

\subsection{Overview}
\label{sec:overview}

\paragraph{Preliminary.} Recently, Mask3D~\cite{Schult23mask3d} and SPFormer~\cite{sun2022superpoint} present a fully end-to-end pipeline, which allows the object queries to directly output instance predictions. With transformer decoders, a fixed number of object queries aggregate information from the global features (either multi-scale voxel features~\cite{Schult23mask3d} or superpoint features~\cite{sun2022superpoint}) extracted with the backbone. Moreover, similar to Mask2Former~\cite{cheng2021per,cheng2022masked}, they adopt mask attention and rely on the instance masks to guide the cross-attention. Specifically, the cross-attention is masked with the instance masks predicted in the last decoder layer, so that the queries only need to consider the masked features. However, as shown in Fig.~\ref{fig:comp_recall}, the recall of initial instance masks is low in the early stages of training. It hinders the ability to achieve a high-quality result in the subsequent layers and thus increases training difficulty.

\vspace{-0.3cm}
\paragraph{Ours.} Instead of relying on mask attention, we propose an auxiliary center regression task to guide instance segmentation. The overview of our method is shown in Fig.~\ref{fig:overview} (a). We first yield the global positions $\mathcal{P} \in \mathbb{R}^{N\times 3}$ from the input point cloud and extract the global features $\mathcal{F} \in \mathbb{R}^{N\times d}$ using the backbone ($\mathcal{P}$ and $\mathcal{F}$ can be either voxels~\cite{Schult23mask3d} or superpoints~\cite{sun2022superpoint} positions and features). In contrast to existing works, besides the content queries $\mathcal{Q}^{c}_{0} \in \mathbb{R}^{n\times d}$, we also maintain a fixed number of position queries $\mathcal{Q}^{p}_{0} \in [0,1]^{n\times 3}$ that represent the normalized instance centers. $\mathcal{Q}^{p}_{0}$ is randomly initialized and $\mathcal{Q}^{c}_{0}$ is initialized with zero. Given the global positions $\mathcal{P}$ and global features $\mathcal{F}$, our goal is to let the positional queries guide their corresponding content queries in cross-attention, and then iteratively refine both sets of queries, and finally predict the instance centers, classes and masks. For the $t$-th decoder layer, this process is formulated as
\begin{equation}
    \mathbf{Center}_{t} = \mathbf{MLP}_{center}(\mathbf{Q}^{c}_{t}) + \mathbf{Q}^{p}_{t-1},
\end{equation}
\begin{equation}
    \mathbf{Class}_{t} = \mathbf{MLP}_{cls}(\mathbf{Q}^{c}_{t}),
\end{equation}
\begin{equation}
    \mathbf{Mask}_{t} = \sigma(\mathbf{Q}^{c}_{t}\cdot {\mathcal{F}_{mask}}^{T}) < 0.5, \; \mathcal{F}_{mask}=\mathbf{MLP}_{mask}(\mathcal{F}),
\end{equation}
where $\mathbf{Center}_{t} \in \mathbb{R}^{n\times 3}$, $\mathbf{Class}_{t} \in \mathbb{R}^{n\times K}$ and $\mathbf{Mask}_{t} \in \{0,1\}^{n\times N}$ are the predicted centers, classification logits, and the instance masks.

\subsection{Position-aware Designs}
\label{sec:query_refinement_module}

To effectively support the center regression task and improve the recall of initial instance masks, we propose a series of position-aware designs as follows.

\paragraph{Learnable Position Query.}

Unlike previous works~\cite{Schult23mask3d,sun2022superpoint}, we introduce an additional set of position queries $\mathcal{Q}^{p}_{0} \in [0,1]^{n\times 3}$. Since the range of points varies significantly among different scenes, the initial position queries are stored in a normalized form as learnable parameters followed by sigmoid function. Basically, we can obtain the absolute positions 
$\hat{\mathcal{Q}}^{p}_{t} \in \mathbb{R}^{n\times 3}$ from the normalized position queries $\mathcal{Q}^{p}_{t}$ for a given input scene as

\begin{equation}
\hat{\mathcal{Q}}^{p}_{t} = \mathcal{Q}^{p}_{t} \cdot (p_{max} - p_{min}) + p_{min},
\end{equation}
where the $p_{min}, p_{max} \in \mathbb{R}^{3}$ represent the minimum and maximum coordinates of the input scene, respectively. The resultant $\hat{\mathcal{Q}}^{p}_{t}$ explicitly represents the positions of the corresponding content queries $\mathcal{Q}^{c}_{t}$.

It is notable that the initial position queries are densely spread throughout the 3D space. Also, every query aggregates features from its local region. This design choice makes it easier for the initial queries to capture the objects in a scene with a high recall, as shown in Fig.~\ref{fig:comp_recall}. It overcomes the low-recall issue caused by initial instance masks, and consequently reduces the training complexity of the subsequent layers.



\begin{figure}[t]
\begin{center}
\includegraphics[width=1.0\linewidth]{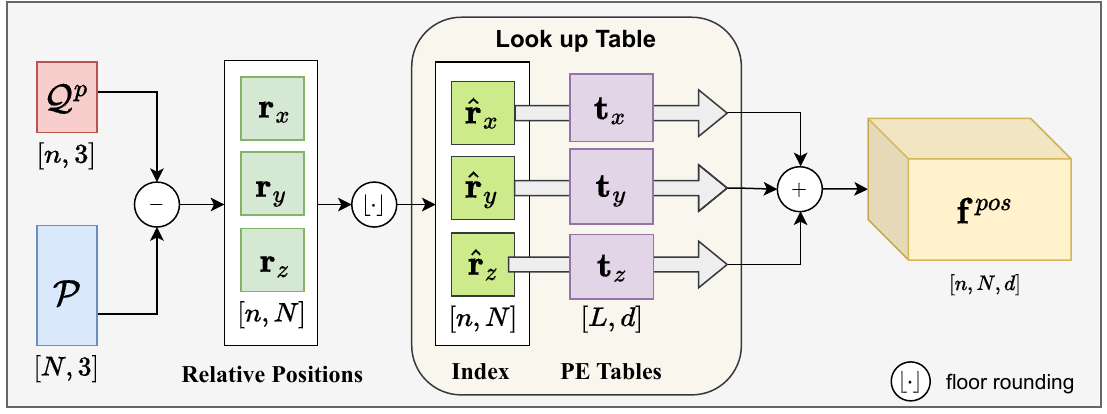}
\end{center}
\vspace{-0.3cm}
\caption{The illustration for relative position encoding. We note the shape of the corresponding tensor at the bottom.}
\label{fig:rpe}
\end{figure}

\paragraph{Relative Position Encoding.}

Other than the absolute position encoding (\eg, Fourier or sine transformations), we also adopt contextual relative position encoding in cross-attention. Inspired by \cite{lai2022stratified}, we first calculate the relative positions $\mathbf{r} \in \mathbb{R}^{n\times N \times 3}$ between the position queries $\hat{\mathcal{Q}}^{p}_{t} \in \mathbb{R}^{n\times 3}$ and the global positions $\mathcal{P} \in \mathbb{R}^{N \times 3}$, and quantize it into discrete integers $\mathbf{\hat{r}} \in \mathbb{Z}^{n\times N \times 3}$ as shown in Fig.~\ref{fig:rpe}. It is formulated as
\begin{equation}
\vspace{-0.1cm}
    \mathbf{r} = \hat{\mathcal{Q}}^{p}_{t} - \mathcal{P}, \quad {\mathbf{\hat{r}}} = \lfloor \frac{\mathbf{r}}{s} \rfloor + \frac{L}{2},
\vspace{-0.1cm}
\end{equation}
where $s$ denotes the quantization size, $L$ denotes the length of position encoding table. We plus $\frac{L}{2}$ to ensure the discrete relative positions are non-negative.

Then, we use the discrete relative positions $\mathbf{\hat{r}}$ as indices to look up the corresponding position encoding tables $\mathbf{t} \in \mathbb{R}^{3\times L\times d}$, as illustrated in Fig.~\ref{fig:rpe}. Formally, the relative position encoding $\mathbf{f}^{pos} \in \mathbb{R}^{n\times N \times d}$ is yielded as
\begin{equation}
    \mathbf{f}^{pos} = \mathbf{t}[0, \mathbf{\hat{r}}_x] + \mathbf{t}[1, \mathbf{\hat{r}}_y] + \mathbf{t}[2, \mathbf{\hat{r}}_z],
\end{equation}
where $\mathbf{\hat{r}}_x, \mathbf{\hat{r}}_y, \mathbf{\hat{r}}_z \in \mathbb{Z}^{n\times N}$ are the discrete relative positions along the $x$, $y$, and $z$-axis, respectively.

Further, the relative position encoding $\mathbf{f}^{pos}$ performs dot product with the query features $\mathbf{f}^q \in \mathbb{R}^{n\times d}$ or key features $\mathbf{f}^k \in \mathbb{R}^{N\times d}$ in the cross-attention, which is formulated as 
\begin{equation}
    \mathbf{pos\_bias}_{i,j} = \mathbf{f}^{pos}_{i,j} \cdot \mathbf{f}^q_{i} + \mathbf{f}^{pos}_{i,j} \cdot \mathbf{f}^k_{j},
\end{equation}
where $\mathbf{pos\_bias} \in \mathbb{R}^{n\times N}$ is the positional bias. It is then added to the cross-attention weights, followed by the softmax function, as shown in Fig.~\ref{fig:overview} (b). 

It is worth noting that the RPE offers a greater degree of flexibility and error-insensitivity, compared to mask attention. In essence, RPE can be likened to a soft mask that has the ability to adjust attention weights flexibly, instead of hard masking. Another advantage of RPE is that it integrates semantic information (\eg, object size and class) and thus can harvest local information selectively. This is accomplished by the interaction between the relative positions and the semantic features (\ie, $\mathbf{f}^q$ and $\mathbf{f}^k$).

\paragraph{Iterative Refinement.}

Since the content queries in our decoder layers are updated regularly, it is not optimal to maintain frozen position queries throughout the decoding process. Additionally, the initial position queries are static, so it is beneficial to adapt them to the specific input scene in the subsequent layers. To that end, we iteratively refine the position queries based on the content queries. Specifically, as shown in Fig.~\ref{fig:overview} (b), we leverage an MLP to predict a center offset $\Delta p_t$ from the updated content query $\mathcal{Q}^c_{t+1}$. We then add it to the previous position query $\hat{\mathcal{Q}}^p_{t}$ as
\begin{align}
\begin{aligned}
    \Delta p_{t} \; & = \textbf{MLP}_{center}(\mathcal{Q}^c_{t+1}), \\
    \hat{\mathcal{Q}}^p_{t+1} & = \hat{\mathcal{Q}}^p_{t} + \Delta p_t.
\end{aligned}
\end{align}

\paragraph{Center Matching \& Loss.} 
To eliminate the need for duplicate removal methods such as non-maximum suppression (NMS), bipartite matching is adopted during training. Existing works~\cite{Schult23mask3d,sun2022superpoint} rely on semantic predictions and binary masks to match the ground truths.

In contrast, to support center regression, we also incorporate center distance in bipartite matching. Since we require the queries to only attend to a local region, it is critical to ensure that they only match with nearby ground-truth objects. To achieve this, we adapt the matching costs formulation as follows
\begin{align}
\begin{aligned}
    \mathcal{C}_{cls}&(k, \hat{k}) = \operatorname{CE}(\mathbf{Class}_k, \hat{k}), \\
    \mathcal{C}_{dice}&(k, \hat{k}) = \operatorname{DICE}(\mathbf{Mask}_k, \mathbf{Mask}_{\hat{k}}), \\
    \mathcal{C}_{bce}&(k, \hat{k}) = \operatorname{BCE}(\mathbf{Mask}_k, \mathbf{Mask}_{\hat{k}}), \\
    \mathcal{C}_{center}&(k, \hat{k}) = \operatorname{L_1}(\mathbf{Center}_k, \mathbf{Center}_{\hat{k}}),\\
    \mathcal{C}&(k, \hat{k}) = \lambda_{cls}\mathcal{C}_{cls}(k, \hat{k}) + \lambda_{dice}\mathcal{C}_{dice}(k, \hat{k}) \\
    & + \lambda_{bce}\mathcal{C}_{bce}(k, \hat{k}) + \lambda_{center}\mathcal{C}_{center}(k, \hat{k}),
\end{aligned}
\end{align}
where $k$ and $\hat{k}$ denotes a predicted and ground-truth instance, respectively, $\mathcal{C} \in \mathbb{R}^{n \times n_{inst}}$ denotes the matching cost matrix, and $\lambda$ denotes the cost weights.

The Hungarian Algorithm is then applied on $\mathcal{C}$ to yield the one-to-one matching result $\hat{\sigma} \in \mathbb{Z}^{n}$, which is followed by the loss function as
\begin{align}
\begin{aligned}
    \hat{\sigma} = & \arg\min_{\sigma: \sigma_i\neq \sigma_j, \forall i\neq j} \sum_{i=1}^{n} \mathcal{C}(i, \sigma_i), \\ 
    \mathcal{L} = & \lambda_{cls} \operatorname{CE}(\mathbf{Class}_i, \hat{\sigma_i}) \\ 
    & + \lambda_{dice} \operatorname{DICE}(\mathbf{Mask}_i, \mathbf{Mask}_{\hat{\sigma_i}}) \\ 
    & + \lambda_{bce} \operatorname{BCE}(\mathbf{Mask}_i, \mathbf{Mask}_{\hat{\sigma_i}})  \\
    & + \lambda_{center} \operatorname{L_{1}}(\mathbf{Center}_i, \mathbf{Center}_{\hat{\sigma_i}}). 
\end{aligned}
\end{align}

\section{Experiment}

This section first provides an overview of the experimental setup in Sec.~\ref{sec:exp_setting}.  We then present the 3D instance segmentation results in Sec.~\ref{sec:exp_inst_seg}. Additionally, we conduct an extensive ablation study in Sec.~\ref{sec:exp_ablation}. Furthermore, we showcase the object detection results and visual comparisons in Sections~\ref{sec:exp_det} and \ref{sec:visual_comp}, respectively. Code and models will be made publicly available.

\subsection{Experimental Setting}
\label{sec:exp_setting}

\paragraph{Network Architecture.}

For both ScanNetv2~\cite{dai2017scannet} and ScanNet200~\cite{rozenberszki2022language}, we follow previous works~\cite{sun2022superpoint, jiang2020pointgroup, vu2022softgroup, chen2021hierarchical, liang2021instance} to use 5-layer U-Net as the backbone. The initial channel is set to 32. Unless otherwise specified, we use the coordinates and colors as the input features. We use 6 layers of Transformer decoders, where the head number is set to 8 and the hidden and feed-forward dimensions are set to 256 and 1024, respectively. We adopt Fourier absolute position encoding with the temperature set to 10,000. The quantization size for RPE is set to 0.1m, and the length of the RPE table is 48. Unless otherwise specified, we choose \cite{sun2022superpoint} as the baseline model, since it has achieved the best performance on ScanNetv2 \textit{val} set so far. For the S3DIS~\cite{armeni_cvpr16} dataset, following Mask3D~\cite{Schult23mask3d}, we use Res16UNet34C~\cite{choy20194d} as the backbone and employ 4 decoders to attend to the coarsest four scales, and this is repeated 3 times with the shared parameters. The decoder hidden and feed-forward dimensions are set to 128 and 1024, respectively. 

\paragraph{Datasets.}

We use the ScanNetv2~\cite{dai2017scannet}, ScanNet200~\cite{rozenberszki2022language} and S3DIS~\cite{armeni_cvpr16} datasets for evaluation. All of them are challenging large-scale indoor scene datasets. 

The ScanNetv2 dataset comprises 1201 scenes for training, and an additional 312 and 100 indoor scenes for validation and testing, respectively. The scenes are captured with RGB-D cameras and annotated with 20 semantic labels, 18 of which are instance classes. The ScanNet200 dataset adopts the same point cloud data, but it offers more diverse annotations, covering 200 classes, 198 of which are instance classes.

The S3DIS dataset contains 271 rooms in 6 areas of three buildings, and 13 semantic categories are annotated. Following previous works, the scenes in Area 5 are used for validation and the others are for training. 

\paragraph{Implementation Details.}
We adopt one RTX 3090 GPU for training on ScanNet and ScanNet200, and one A100 GPU on S3DIS. Following previous works, we use AdamW~\cite{loshchilov2017decoupled} optimizer with the learning rate and weight decay set to 0.0001 and 0.05, respectively. We adopt poly scheduler on ScanNet and ScanNet200, and onecycle scheduler on S3DIS. The batch size is set to 4. For the weights of matching costs and losses, ($\lambda_{cls}$, $\lambda_{bce}$, $\lambda_{dice}$, $\lambda_{center}$) are set to (0.5, 1.0, 1.0, 0.5) on ScanNet and ScanNet200, and (2.0, 5.0, 1.0, 0.5) on S3DIS. The voxel size is set to 0.02m. We limit the points number up to 250,000. Otherwise, we crop the scene by cubic windows iteratively until the point number is lower than the limit. During inference, we select the top 100 instances with the highest scores and set the minimum points number per instance to 100.

\begin{table*}[t]
    \centering
    \tabcolsep=0.13cm
    {
        \begin{footnotesize}
        \begin{tabular}{ l | c c | c c c c c c c c c c c c c c c c c c}
            \toprule
            
            Method & mAP & mAP\textsubscript{50} & \rotatebox{90}{bath} & \rotatebox{90}{bed} & \rotatebox{90}{bkshf} & \rotatebox{90}{cabinet} & \rotatebox{90}{chair} & \rotatebox{90}{counter} & \rotatebox{90}{curtain} & \rotatebox{90}{desk} & \rotatebox{90}{door} & \rotatebox{90}{other} & \rotatebox{90}{picture} & \rotatebox{90}{fridge} & \rotatebox{90}{s. cur.} & \rotatebox{90}{sink} & \rotatebox{90}{sofa} & \rotatebox{90}{table} & \rotatebox{90}{toilet} & \rotatebox{90}{wind.} \\ 
            
            \specialrule{0em}{0pt}{1pt}
            \hline
            \specialrule{0em}{0pt}{1pt}
            
            3D-BoNet~\cite{YangWCHWMT19} & 25.3 & 48.8 & 51.9 & 32.4 & 25.1 & 13.7 & 34.5 & 3.1 & 41.9 & 6.9 & 16.2 & 13.1 & 5.2 & 20.2 & 33.8 & 14.7 & 30.1 & 30.3 & 65.1 & 17.8 \\

            MTML~\cite{lahoud20193d} & 28.2 & 40.2 & 57.7 & 38.0 & 18.2 & 10.7 & 43.0 & 0.1 & 42.2 & 5.7 & 17.9 & 16.2 & 7.0 & 22.9 & 51.1 & 16.1 & 49.1 & 31.3 & 65.0 & 16.2 \\

            GICN~\cite{liu2020learning} & 34.1 & 63.8 & 58.0 & 37.1 & 34.4 & 19.8 & 46.9 & 5.2 & 56.4 & 9.3 & 21.2 & 21.2 & 12.7 & 34.7 & 53.7 & 20.6 & 52.5 & 32.9 & 72.9 & 24.1 \\

            3D-MPA~\cite{engelmann20203d} & 35.5 & 61.1 & 45.7 & 48.4 & 29.9 & 27.7 & 59.1 & 4.7 & 33.2 & 21.2 & 21.7 & 27.8 & 19.3 & 41.3 & 41.0 & 19.5 & 57.4 & 35.2 & 84.9 & 21.3 \\

            Dyco3D~\cite{he2021dyco3d} & 39.5 & 64.1 & 64.2 & 51.8 & 44.7 & 25.9 & 66.6 & 5.0 & 25.1 & 16.6 & 23.1 & 36.2 & 23.2 & 33.1 & 53.5 & 22.9 & 58.7 & 43.8 & 85.0 & 31.7 \\

            PE~\cite{zhang2021point} & 39.6 & 64.5 & 66.7 & 46.7 & 44.6 & 24.3 & 62.4 & 2.2 & 57.7 & 10.6 & 21.9 & 34.0 & 23.9 & 48.7 & 47.5 & 22.5 & 54.1 & 35.0 & 81.8 & 27.3 \\

            PointGroup~\cite{jiang2020pointgroup} & 40.7 & 63.6 & 63.9 & 49.6 & 41.5 & 24.3 & 64.5 & 2.1 & 57.0 & 11.4 & 21.1 & 35.9 & 21.7 & 42.8 & 66.6 & 25.6 & 56.2 & 34.1 & 86.0 & 29.1 \\

            HAIS~\cite{chen2021hierarchical} & 45.7 & 69.9 & 70.4 & 56.1 & 45.7 & 36.4 & 67.3 & 4.6 & 54.7 & 19.4 & 30.8 & 42.6 & 28.8 & 45.4 & 71.1 & 26.2 & 56.3 & 43.4 & 88.9 & 34.4 \\

            OccuSeg~\cite{han2020occuseg} & 48.6 & 67.2 & 80.2 & 53.6 & 42.8 & 36.9 & 70.2 & 20.5 & 33.1 & 30.1 & 37.9 & 47.4 & 32.7 & 43.7 & 86.2 & 48.5 & 60.1 & 39.4 & 84.6 & 27.3 \\
            
            SoftGroup~\cite{vu2022softgroup} & 50.4 & 76.1 & 66.7 & 57.9 & 37.2 & 38.1 & 69.4 & 7.2 & 67.7 & 30.3 & 38.7 & 53.1 & 31.9 & 58.2 & 75.4 & 31.8 & 64.3 & 49.2 & 90.7 & 38.8 \\

            SSTNet~\cite{liang2021instance} & 50.6 & 69.8 & 73.8 & 54.9 & 49.7 & 31.6 & 69.3 & 17.8 & 37.7 & 19.8 & 33.0 & 46.3 & 57.6 & 51.5 & 85.7 & 49.4 & 63.7 & 45.7 & 94.3 & 29.0 \\

            SPFormer~\cite{sun2022superpoint} & 54.9 & 77.0 & 74.5 & 64.0 & 48.4 & 39.5 & 73.9 & 31.1 & 56.6 & 33.5 & 46.8 & 49.2 & 55.5 & 47.8 & 74.7 & 43.6 & 71.2 & 54.0 & 89.3 & 34.3 \\
            
            Mask3D$^*$~\cite{Schult23mask3d} & 56.6 & 78.0 & 92.6 & 59.7 & 40.8 & 42.0 & 73.7 & 23.9 & 59.8 & 38.6 & 45.8 & 54.9 & 56.8 & 71.6 & 60.1 & 48.0 & 64.6 & 57.5 & 92.2 & 36.4 \\
            
            \specialrule{0em}{0pt}{1pt}
            \hline
            \specialrule{0em}{0pt}{1pt}
            
            Ours & 57.8 & 77.4 & 77.8 & 64.9 & 52.0 & 44.9 & 76.1 & 25.3 & 58.4 & 39.1 & 53.0 & 47.2 & 61.7 & 49.9 & 79.5 & 47.3 & 74.5 & 54.8 & 96.0 & 37.4 \\

            Ours$^\ddagger$ & \textbf{59.6} & \textbf{78.6} & 88.9 & 72.1 & 44.8 & 46.0 & 76.8 & 25.1 & 55.8 & 40.8 & 50.4 & 53.9 & 61.6 & 61.8 & 85.8 & 48.2 & 68.4 & 55.1 & 93.1 & 45.0 \\
            
            \bottomrule                                   
        \end{tabular}
        \end{footnotesize}
    }
    \caption{3D instance segmentation results on ScanNet \textit{test} set. $\;\;^*$ denotes using Res16UNet34C (twice as many parameters as ours) as the backbone. $\;\;^{\ddagger}$ denotes using surface normal. Methods published before the submission deadline (03/08/2023) are listed.}
    \label{tab:exp_scannet_test}   
\end{table*}

\begin{table}[t]
    \centering
    \tabcolsep=0.7cm
    {
        \begin{tabular}{ l | c c }
            \toprule
            
            Method & mAP & mAP\textsubscript{50}\\ 
            
            \specialrule{0em}{0pt}{1pt}
            \hline
            \specialrule{0em}{0pt}{1pt}
            
            GSPN~\cite{yi2019gspn} & 19.3 & 37.8  \\

            MTML~\cite{lahoud20193d} & 20.3 & 40.2 \\

            3D-MPA~\cite{engelmann20203d} & 35.5 & 59.1 \\

            Dyco3D~\cite{he2021dyco3d} & 35.4 & 57.6 \\
            
            PointGroup~\cite{jiang2020pointgroup} & 34.8 & 56.7 \\

            MaskGroup~\cite{zhong2022maskgroup} & 42.0 & 63.3 \\

            HAIS~\cite{chen2021hierarchical} & 43.5 & 64.1 \\

            OccuSeg~\cite{han2020occuseg} & 44.2 & 60.7 \\
            
            SoftGroup~\cite{vu2022softgroup} & 46.0 & 67.6 \\

            SSTNet~\cite{liang2021instance} & 49.4 & 64.3 \\

            SPFormer~\cite{sun2022superpoint} & 56.3 & 73.9 \\
            
            Mask3D$^*$~\cite{Schult23mask3d} & 55.2 & 73.7 \\
            
            \specialrule{0em}{0pt}{1pt}
            \hline
            \specialrule{0em}{0pt}{1pt}
            
            Ours & 58.4 & 75.9 \\

            Ours$^\ddagger$ & \textbf{59.9} & \textbf{76.5} \\
            
            \bottomrule            
        \end{tabular}
    }
    \vspace{0.3cm}
    \caption{3D instance segmentation results on ScanNet \textit{val} set. $\;\;^*$ denotes using Res16UNet34C (twice as many parameters as ours) as the backbone. $\;\;^{\ddagger}$ denotes using surface normal.}
    \label{tab:exp_scannet_val}   
\end{table}

\begin{table}[t]
    \centering
    \tabcolsep=0.4cm
    {
        \begin{tabular}{ l | c c c c c c }
            \toprule
            
            Method & mAP & mAP\textsubscript{50} & mAP\textsubscript{25} \\ 
            
            \specialrule{0em}{0pt}{1pt}
            \hline
            \specialrule{0em}{0pt}{1pt}
            
            SPFormer$^\dagger$~\cite{sun2022superpoint} & 25.2 &	33.8 & 39.6 \\
            
            Mask3D$^*$~\cite{Schult23mask3d} & 27.4 & 37.0 & 42.3 \\
            
            \specialrule{0em}{0pt}{1pt}
            \hline
            \specialrule{0em}{0pt}{1pt}
            
            Ours & \textbf{29.2} & \textbf{38.2} & \textbf{43.3} \\
            
            \bottomrule                                   
        \end{tabular}
    }
    \vspace{0.3cm}
    \caption{3D instance segmentation results on ScanNet200 \textit{val} set. $\;\;^\dagger$ denotes reproduced results. $\;\;^*$ denotes using Res16UNet34C.}
    \label{tab:exp_scannet200_val}   
\end{table}

\begin{table}[t]
    \centering
    \tabcolsep=0.65cm
    {
        \begin{tabular}{ l | c c c c }
            \toprule
            
            Method & mAP\textsubscript{50} & mAP\textsubscript{25} \\ 
            
            \specialrule{0em}{0pt}{1pt}
            \hline
            \specialrule{0em}{0pt}{1pt}
            



            
            PointGroup~\cite{jiang2020pointgroup} & 57.8 & - \\

            MaskGroup~\cite{zhong2022maskgroup} & 65.0 & - \\


            SoftGroup~\cite{vu2022softgroup} & 66.1 & - \\

            SSTNet~\cite{liang2021instance} & 59.3 & - \\

            SPFormer~\cite{sun2022superpoint} & 66.8 & - \\
            
            Mask3D~\cite{Schult23mask3d} & 68.4 & 75.2 \\
            
            \specialrule{0em}{0pt}{1pt}
            \hline
            \specialrule{0em}{0pt}{1pt}
            
            Ours & \textbf{69.1} & \textbf{75.7} \\

            \bottomrule
        \end{tabular}
    }
    \vspace{0.2cm}
    \caption{3D instance segmentation results on S3DIS Area5.}
    \label{tab:exp_s3dis}   
\end{table}

\subsection{Instance Segmentation Results}
\label{sec:exp_inst_seg}

\paragraph{ScanNetv2.} We present the results of instance segmentation on both the ScanNetv2 \textit{test} and \textit{val} sets in Tables~\ref{tab:exp_scannet_test} and \ref{tab:exp_scannet_val}, respectively. Our method achieves a considerable increase in mAP compared to previous works, suggesting a superior ability to capture fine-grained details and produce high-quality instance segmentation. While Mask3D~\cite{Schult23mask3d} slightly outperforms our model in terms of mAP\textsubscript{50}, it is worth noting that this is potentially due to their use of a stronger backbone (\ie, Res16UNet34C with twice as many parameters as ours) and DBSCAN post-processing. Despite this, our approach produces significantly better performance on the ScanNetv2 \textit{val} set than Mask3D, as seen in Table~\ref{tab:exp_scannet_val}.

\paragraph{ScanNet200.} Table~\ref{tab:exp_scannet200_val} presents our comparison with previous state-of-the-art methods on the \textit{val} set of ScanNet200. Our method achieves a significant improvement in comparison to the other methods. Consistent conclustion is also seen on this challenging dataset. It is important to note that previous works employ mask attention, while our approach does not. This verifies the success of our auxiliary center regression task in replacing mask attention.

\paragraph{S3DIS.} As shown in Table~\ref{tab:exp_s3dis}, our method is evaluated on S3DIS Area5. Our approach outperforms previous works. This consistently shows the superiority of our method.

\subsection{Ablation Study}
\label{sec:exp_ablation}

We conduct an extensive ablation study to verify each component of our method as follows.

\begin{table}[t]
    \centering
    \tabcolsep=0.3cm
    {
        \begin{tabular}{ l l | c c c }
            \toprule
            
            Position & Content & mAP & mAP\textsubscript{50} & mAP\textsubscript{25} \\
            
            \specialrule{0em}{0pt}{1pt}
            \hline
            \specialrule{0em}{0pt}{1pt}
            
            FPS & zero & 57.3 & 74.9 & 84.2 \\

            FPS & learnable & 57.1 & 75.0 & 83.4 \\

            learnable & learnable & 58.1 & 75.4 & 84.3 \\
            
            \specialrule{0em}{0pt}{1pt}
            \hline
            \specialrule{0em}{0pt}{1pt}
            
            learnable & zero & \textbf{58.4} & \textbf{75.9} & \textbf{84.5} \\
            

            
            \bottomrule                                   
        \end{tabular}
    }
    \vspace{0.1cm}
    \caption{Ablation study on different initializations for position and content queries.}
    \label{tab:ablation_learnable_query}   
\end{table}

\begin{table}[t]
    \centering
    \tabcolsep=0.25cm
    {
        \begin{tabular}{ l | c c c }
            \toprule
            
            Position Encoding & mAP & mAP\textsubscript{50} & mAP\textsubscript{25} \\
            
            \specialrule{0em}{0pt}{1pt}
            \hline
            \specialrule{0em}{0pt}{1pt}

            No PE & 0.0 & 0.0 & 0.0 \\
            
            Fourier APE & 57.7 & \textbf{76.0} & 83.8 \\

            Content-conditioned APE & 58.0 & 75.7 & 84.3 \\

            \specialrule{0em}{0pt}{1pt}
            \hline
            \specialrule{0em}{0pt}{1pt}
            
            RPE & \textbf{58.4} & 75.9 & \textbf{84.5} \\
            
            \bottomrule                                   
        \end{tabular}
    }
    \vspace{0.1cm}
    \caption{Ablation study on different position encodings.}
    \label{tab:ablation_pos_enc}
\end{table}

\begin{table}[t]
    \centering
    \tabcolsep=0.14cm
    {
        \begin{footnotesize}
        \begin{tabular}{c | c c c | c c c }
            \toprule
            
            ID & iter. refine & center match & center loss & mAP & mAP\textsubscript{50} & mAP\textsubscript{25} \\
            
            \specialrule{0em}{0pt}{1pt}
            \hline
            \specialrule{0em}{0pt}{1pt}

            1 & & \Checkmark & \Checkmark & 57.5 & 75.3 & 84.0 \\
             
            2 &\Checkmark &  & \Checkmark & 56.7 & 74.8 & 84.1 \\
            
            3 &\Checkmark & \Checkmark &  & 56.8 & 74.6 & 84.5 \\
            
            4 &\Checkmark &  &  & 56.4 & 74.7 & 83.7 \\

            \specialrule{0em}{0pt}{1pt}
            \hline
            \specialrule{0em}{0pt}{1pt}
            
            5 &\Checkmark & \Checkmark & \Checkmark & \textbf{58.4} & \textbf{75.9} & \textbf{84.5} \\
            
            \bottomrule                                   
        \end{tabular}
        \end{footnotesize}
    }
    \vspace{0.1cm}
    \caption{Ablation study on iterative refinement and center matching \& loss.}
    \label{tab:ablation_iter_refine_and_center}
\vspace{-0.2cm}
\end{table}

            
            

            
            
            

\begin{figure*}[h]
\begin{center}
\includegraphics[width=1.0\linewidth]{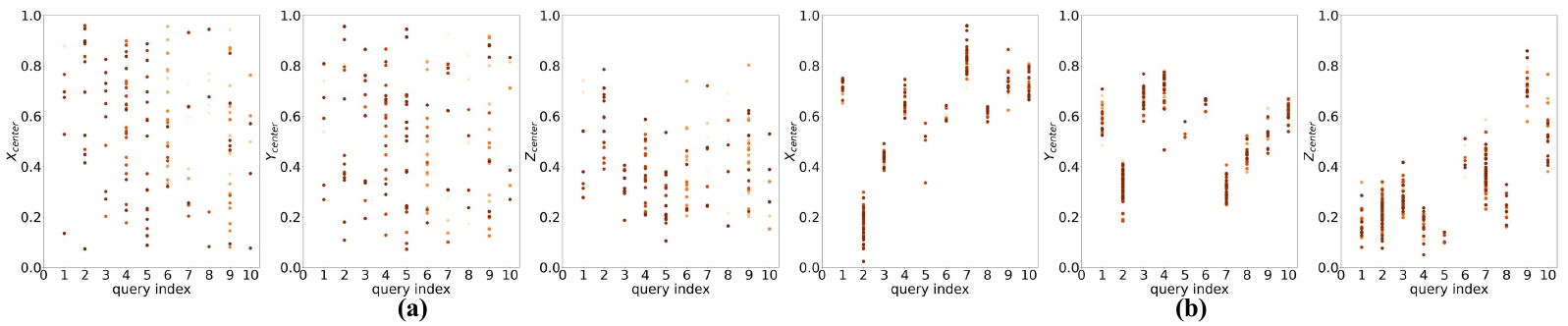}
\end{center}
\vspace{-0.3cm}
\caption{Spatial distribution of ground truths that are matched to the same query at different training iterations. (a): Baseline. (b): Ours. We randomly select 10 object queries and record their matched ground-truth instances in the early training. The three scatter plots for each method represent the center XYZ coordinates of their matched ground-truth instances at different training iterations (the scatter points for a query correspond to different training iterations). More visualization details are given in supplementary material. Our method manifests strong positional matching consistency among different training iterations.}
\label{fig:comp_matching}
\end{figure*}


\paragraph{Learnable Position Query.} The position query aims to provide an explicit center representation to the content query counterpart. Making it learnable intends to learn an optimal initial spatial distribution. We notice that some previous works~\cite{Schult23mask3d,misra2021end} adopt \textit{non-parametric} initial queries, where Furthest Point Sampling (FPS) is used to sample a number of points and transform them into position encodings via Fourier transformation followed by an MLP. We make comparisons in Table~\ref{tab:ablation_learnable_query}. The results show that learnable position query and zero-initialized content query perform best. A potential reason why `FPS' lags behind `learnable' is that the latter learns an optimal spatial distribution. 

Moreover, to show the pattern of the learnable position query, we visualize the spatial distribution of center coordinates of the matched ground truths for a query in Fig.~\ref{fig:comp_matching}. It shows that each query consistently attends to a local region.

\paragraph{Relative Position Encoding.} 

We compare various position encodings that are employed in previous works~\cite{Schult23mask3d,meng2021conditional}, such as Fourier Absolute Position Encoding (APE) and the content query-conditioned APE. Specifically, the latter uses an MLP to project the content query into a $d$-dim diagonal matrix, which then transforms the original absolute position encoding into a new one. It incorporates semantic information into the position encoding but does not consider relative relation. As shown in Table~\ref{tab:ablation_pos_enc}, RPE outperforms the others, which implies that both semantic information and relative relation are beneficial. Also, we notice that if we do not apply any position encoding, the training corrupts. This shows that positional prior is crucial in our framework.

\paragraph{Iterative Refinement.}

We remove the iterative refinement and freeze the position query in all decoder layers, and we find that it causes a performance drop of $0.9\%$ mAP as shown in the first row of Table~\ref{tab:ablation_iter_refine_and_center}. This verifies the effectiveness of iterative refinement. 


\paragraph{Center Matching \& Loss.}

Moreover, to manifest the importance of center matching and center loss, we also conduct ablation studies in Table~\ref{tab:ablation_iter_refine_and_center}. We first remove the center matching and keep the center loss in the second row of the table, and we find that the performance drops by $1.7\%$ mAP. Then we keep the center matching and remove the center loss. The performance also decreases by $1.6\%$ mAP as shown in the 3-rd row. When both are absent, we observe an even larger performance drop ($2.0\%$ mAP) in the 4-th row. The results reveal that both center matching and loss are important to our framework. 



\begin{table}[t]
    \footnotesize
    \centering
    \tabcolsep=0.2cm
    {
        \begin{tabular}{ l | c | c c c }
            \toprule
            
            Method & task & box mAP\textsubscript{50} & box mAP\textsubscript{25} \\ 
            
            \specialrule{0em}{0pt}{1pt}
            \hline
            \specialrule{0em}{0pt}{1pt}

            VoteNet~\cite{qi2019deep} & det & 33.5 & 58.6 \\

            HGNet~\cite{chen2020hierarchical} & det & 34.4 & 61.3 \\

            MLCVNet~\cite{xie2020mlcvnet} & det & 41.4 & 64.5 \\

            GSDN~\cite{gwak2020generative} & det & 34.8 & 62.8 \\

            H3DNet~\cite{zhang2020h3dnet} & det & 48.1 & 67.2 \\

            3DETR~\cite{misra2021end} & det & 47.0 & 65.0 \\

            Group-free~\cite{liu2021group} & det & 52.8 & 69.1 \\

            RBGNet~\cite{wang2022rbgnet} & det & 55.2 & 70.6 \\

            HyperDet3D~\cite{zheng2022hyperdet3d} & det & 57.2 & 70.9 \\            

            FCAF3D~\cite{rukhovich2022fcaf3d} & det & 57.3 & 71.5 \\

            CAGroup3D~\cite{wangcagroup3d} & det & 61.3 & \textbf{75.1} \\
            
            \specialrule{0em}{0pt}{1pt}
            \hline
            \specialrule{0em}{0pt}{1pt}

            3D-MPA~\cite{engelmann20203d} & inst & 49.2 & 64.2 \\
            
            Mask3D~\cite{Schult23mask3d} & inst & 56.2 & 70.2 \\
            

            Ours & inst & \textbf{63.9} & 73.5 & \\

            \bottomrule
        \end{tabular}
    }
    \vspace{0.2cm}
    \caption{3D object detection results on ScanNetv2. For methods designed for the instance segmentation task, the bounding boxes are generated by the instance mask predictions.}
    \label{tab:exp_det}   
\vspace{-0.4cm}
\end{table}

\begin{figure*}[ht]
    \centering

    \begin{minipage}  {0.135\linewidth}
        \centering
        \includegraphics [width=1\linewidth,height=1\linewidth]{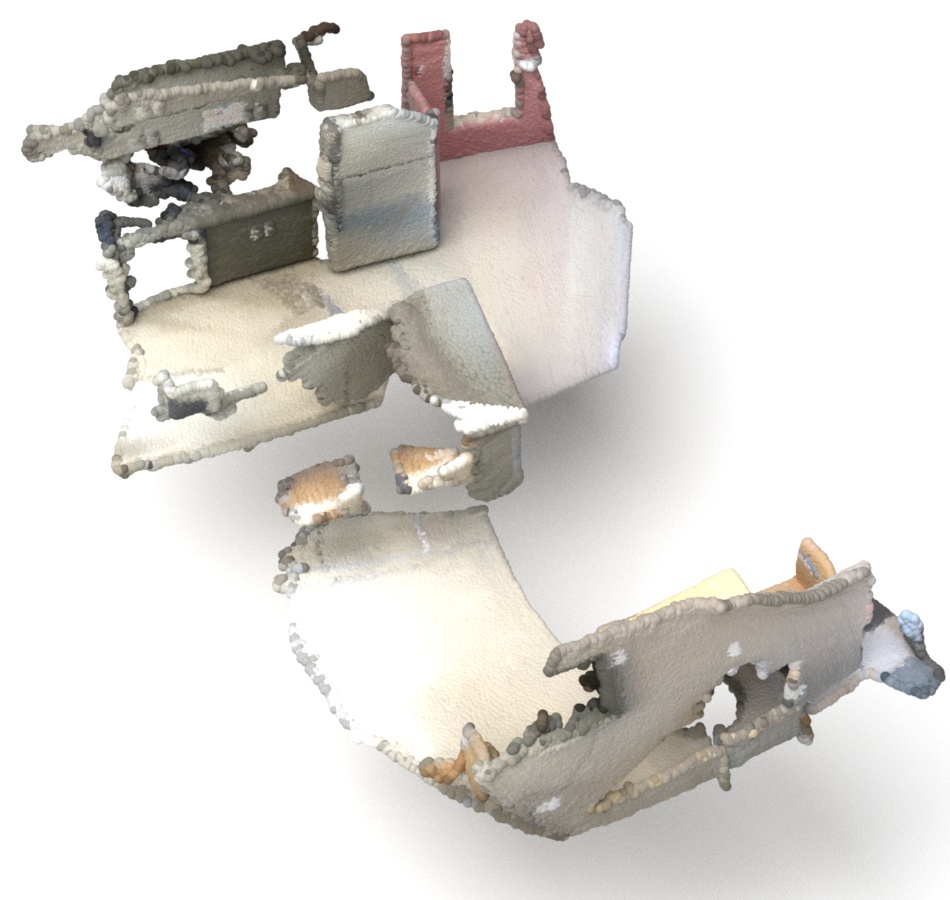}
    \end{minipage}
    \begin{minipage}  {0.135\linewidth}
        \centering
        \includegraphics [width=1\linewidth,height=1\linewidth]{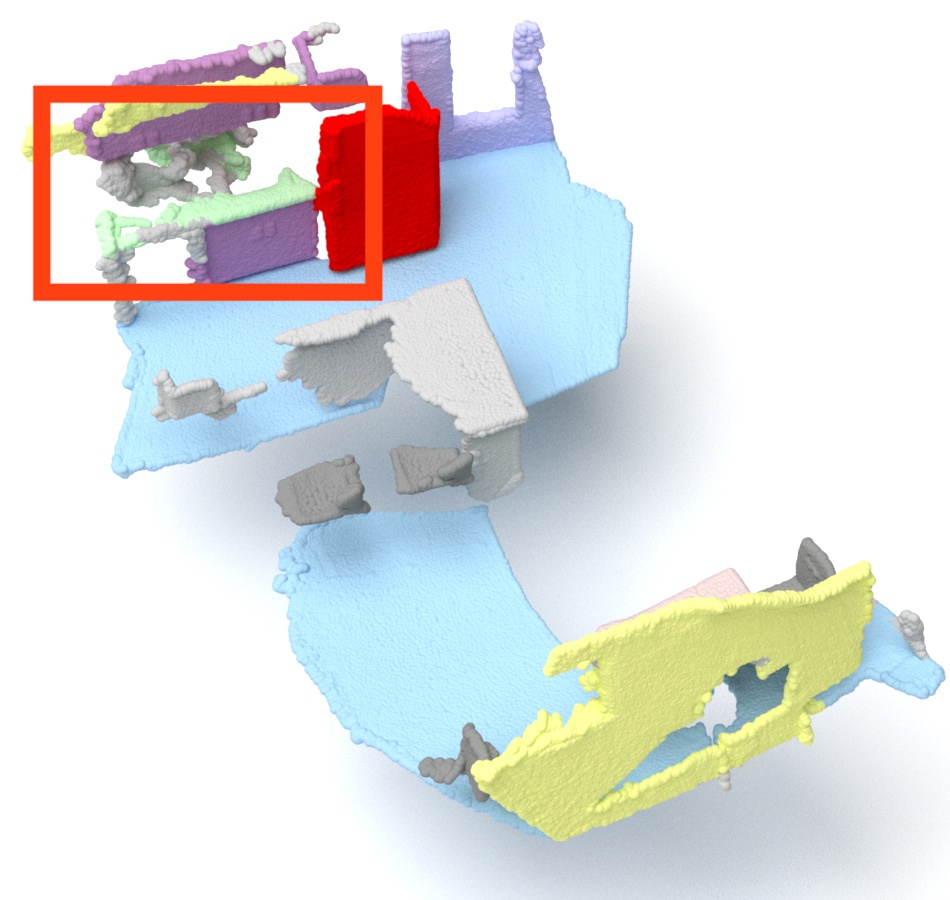}
    \end{minipage}
     \begin{minipage}  {0.135\linewidth}
        \centering
        \includegraphics [width=1\linewidth,height=1\linewidth]{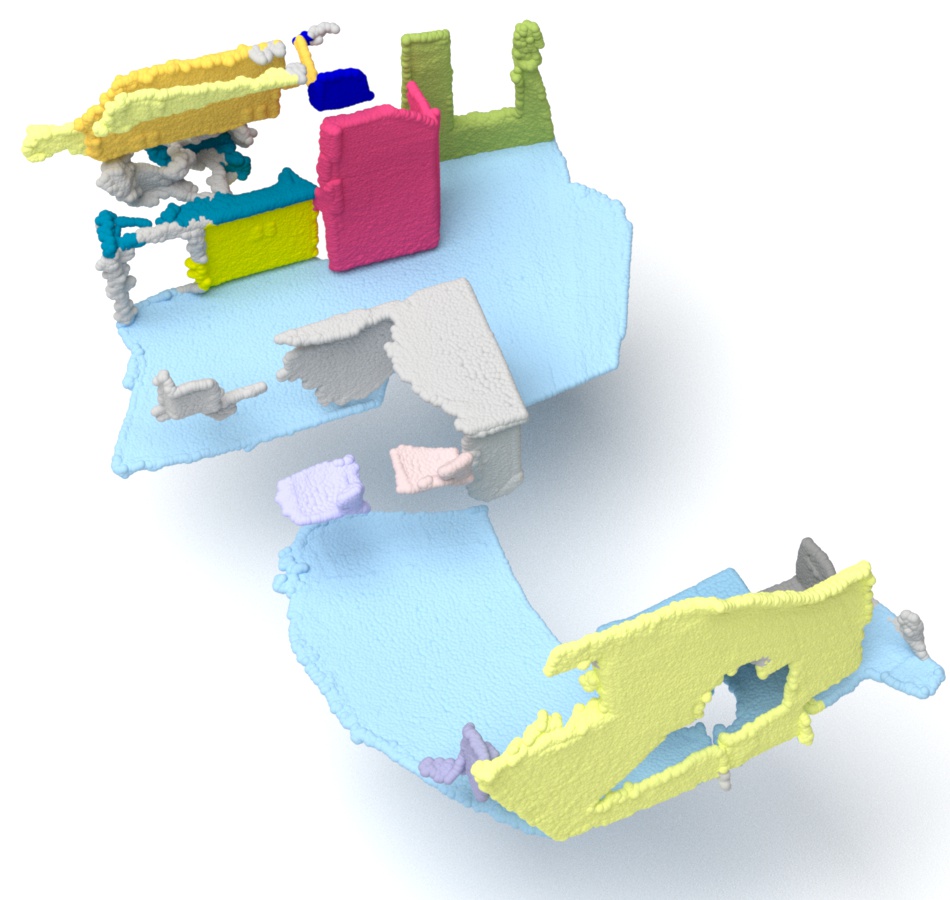}
    \end{minipage} 
    \begin{minipage}  {0.135\linewidth}
        \centering
        \includegraphics [width=1\linewidth,height=1\linewidth]{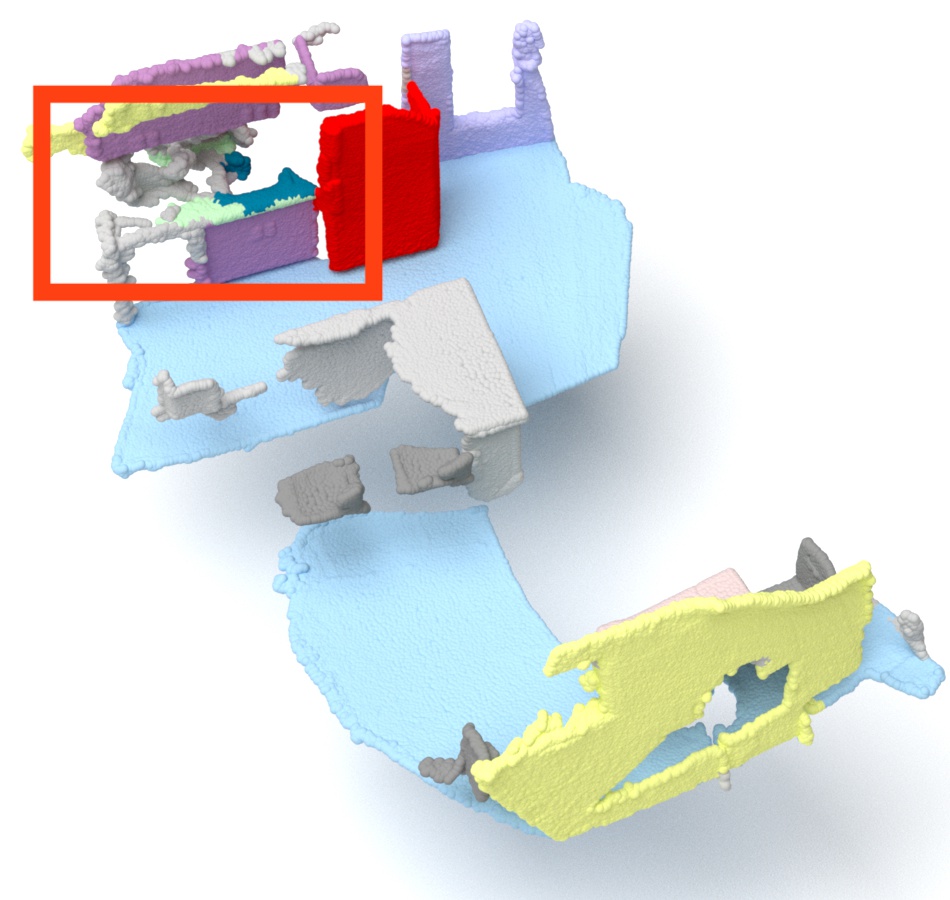}
    \end{minipage}
     \begin{minipage}  {0.135\linewidth}
        \centering
        \includegraphics [width=1\linewidth,height=1\linewidth]{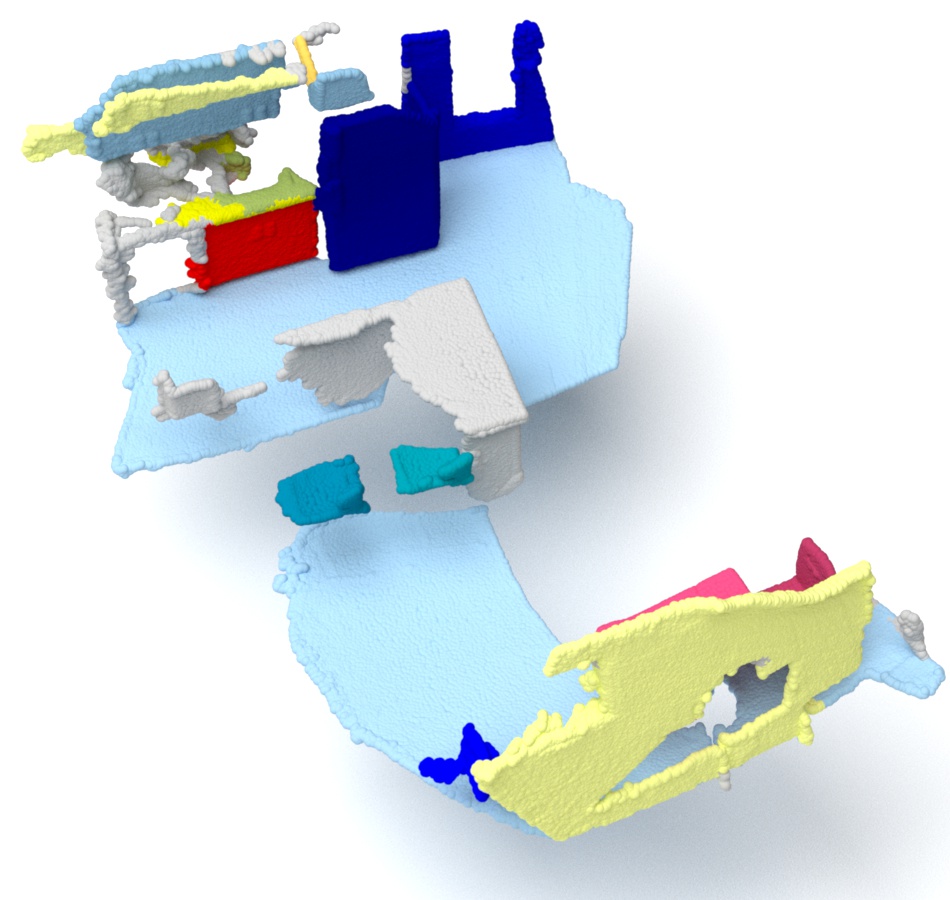}
    \end{minipage} 
     \begin{minipage}  {0.135\linewidth}
        \centering
        \includegraphics [width=1\linewidth,height=1\linewidth]{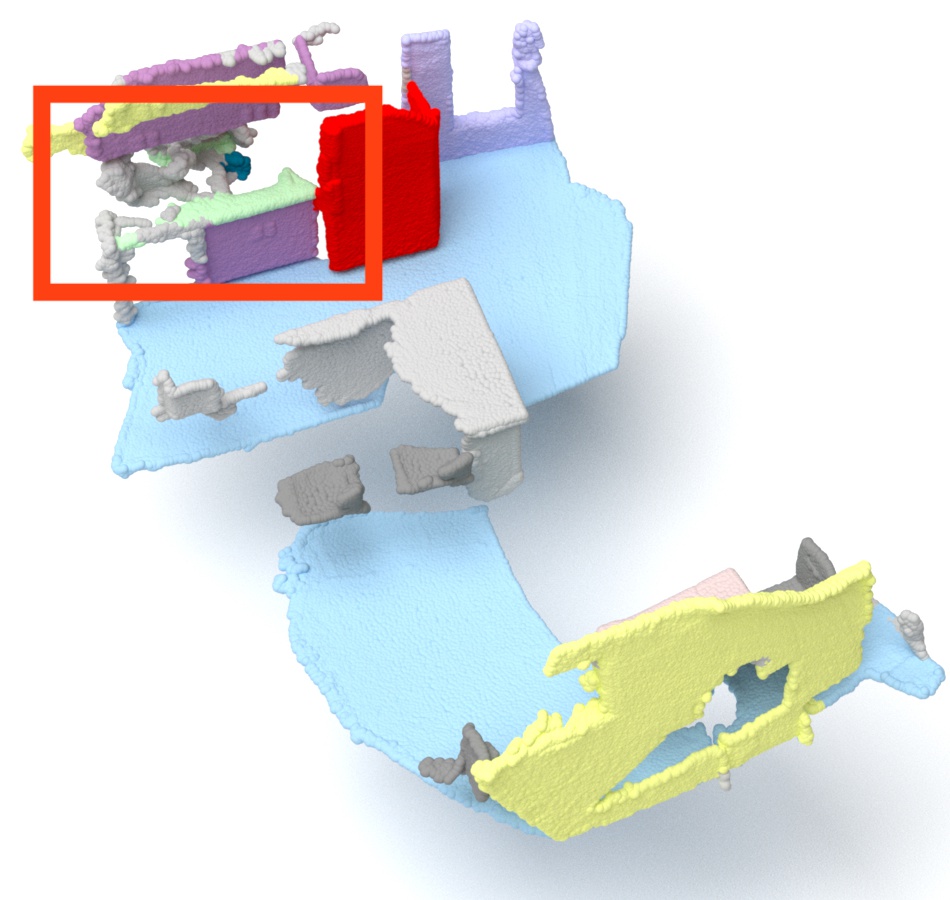}
    \end{minipage} 
     \begin{minipage}  {0.135\linewidth}
        \centering
        \includegraphics [width=1\linewidth,height=1\linewidth]{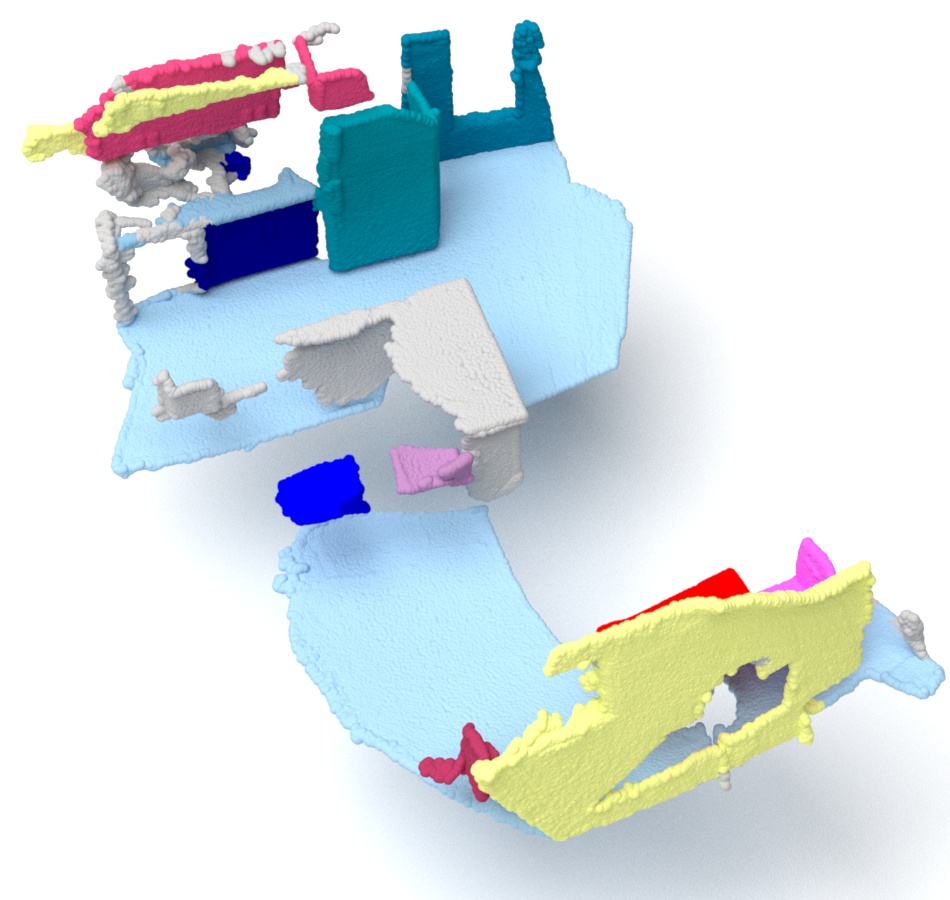}
    \end{minipage} 
  

    \begin{minipage}  {0.135\linewidth}
        \centering
        \includegraphics [width=1\linewidth,height=1\linewidth]{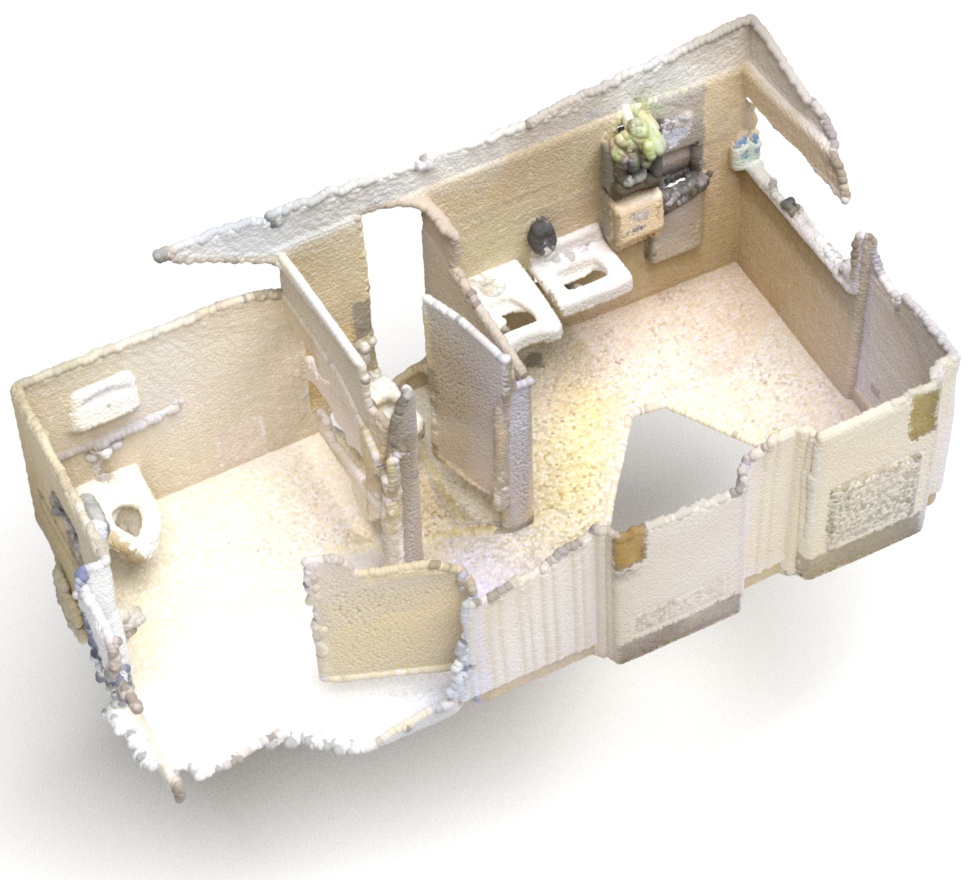}
    \end{minipage}
    \begin{minipage}  {0.135\linewidth}
        \centering
        \includegraphics [width=1\linewidth,height=1\linewidth]{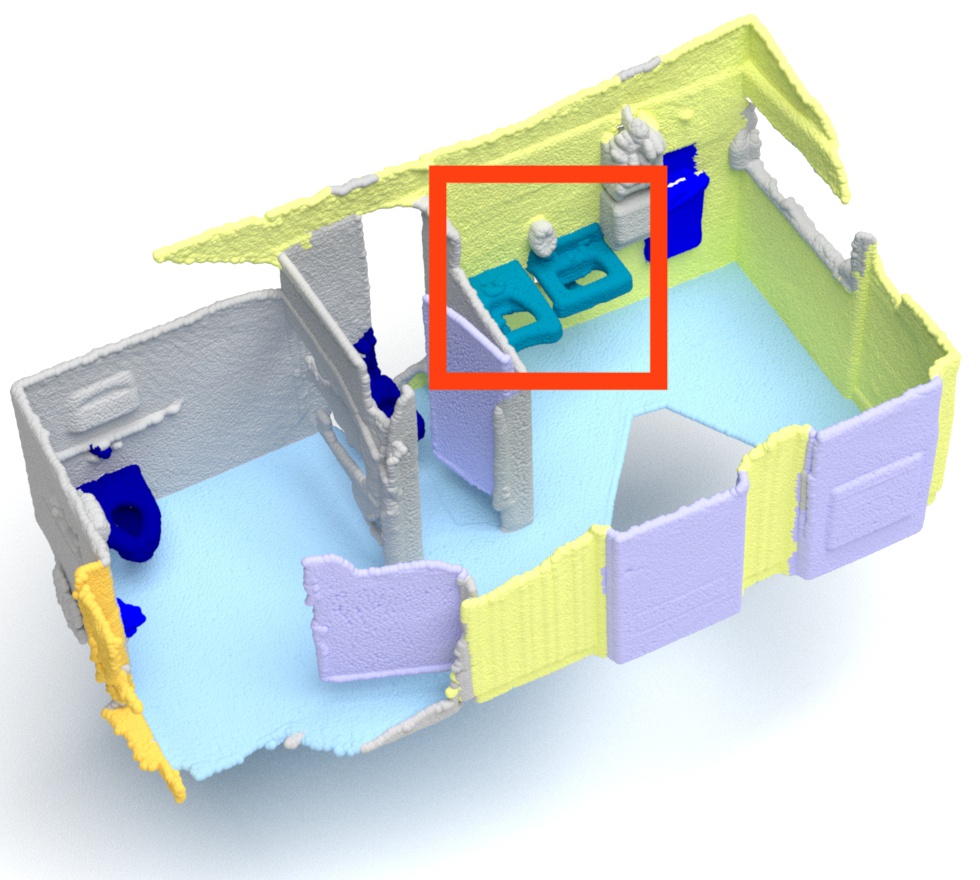}
    \end{minipage}
     \begin{minipage}  {0.135\linewidth}
        \centering
        \includegraphics [width=1\linewidth,height=1\linewidth]{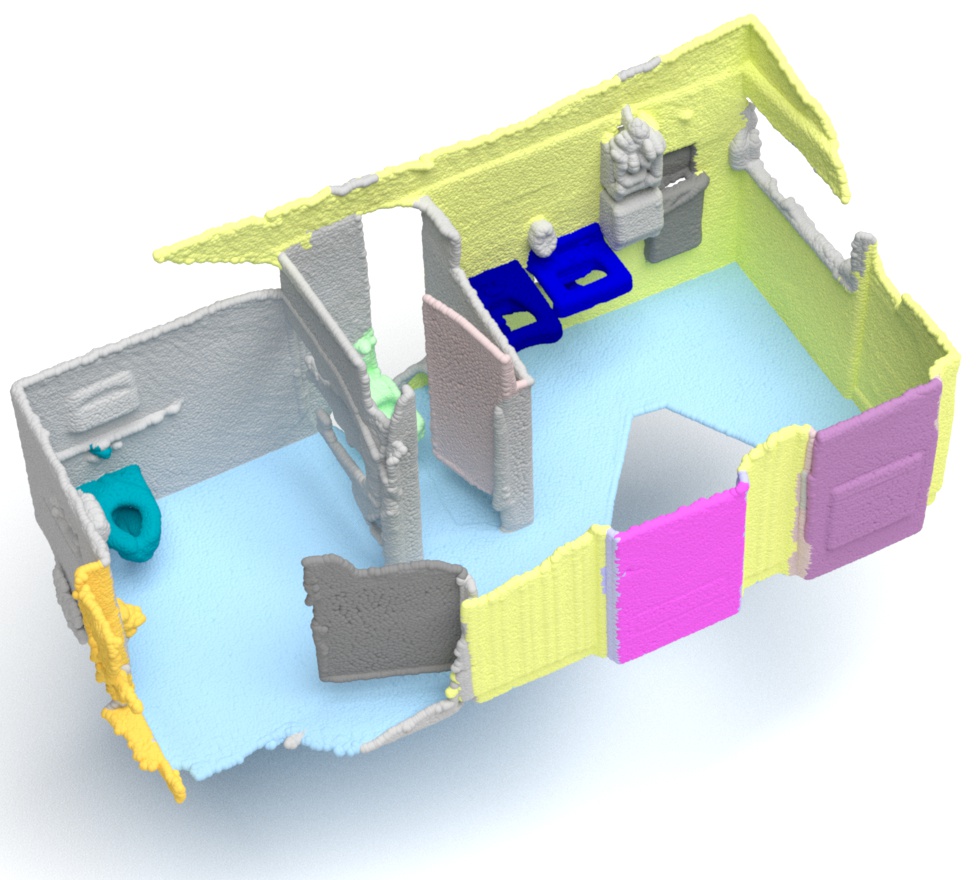}
    \end{minipage} 
    \begin{minipage}  {0.135\linewidth}
        \centering
        \includegraphics [width=1\linewidth,height=1\linewidth]{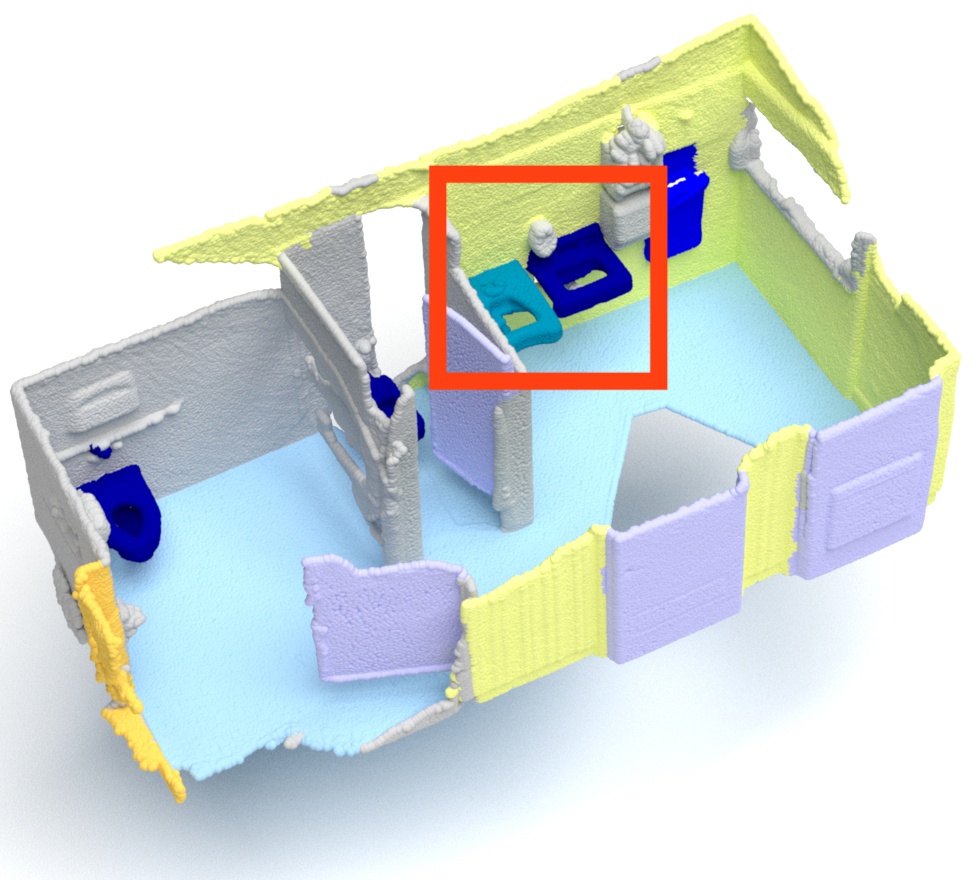}
    \end{minipage}
     \begin{minipage}  {0.135\linewidth}
        \centering
        \includegraphics [width=1\linewidth,height=1\linewidth]{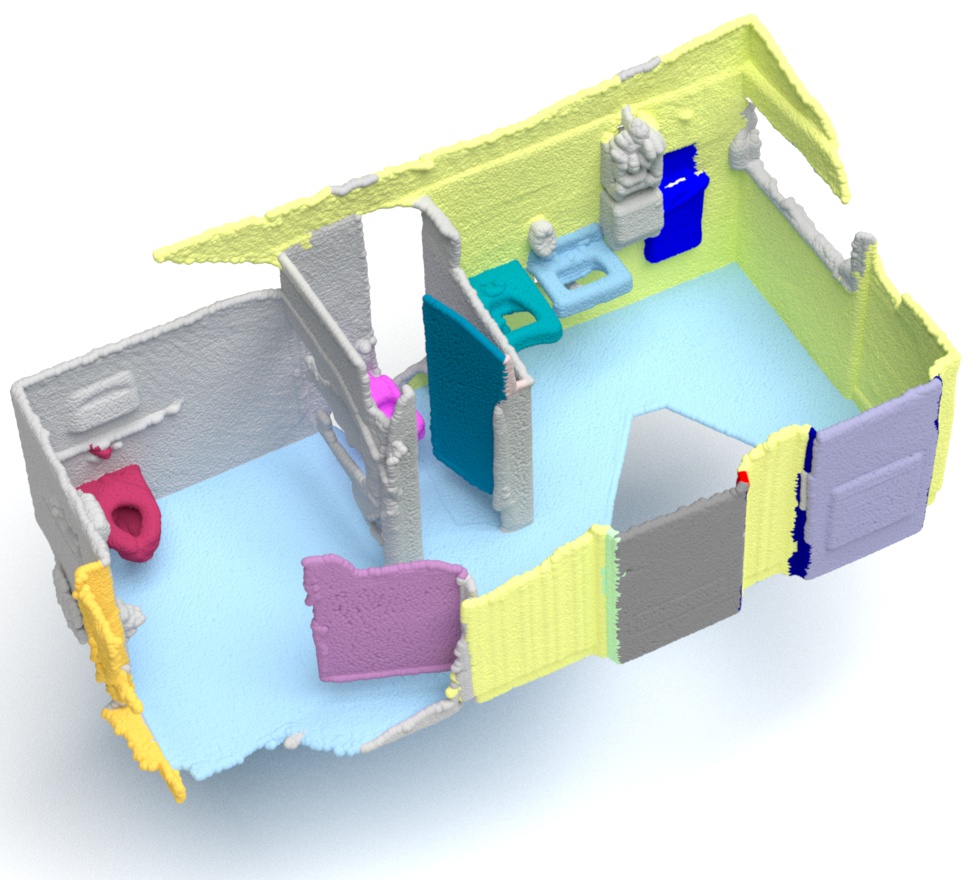}
    \end{minipage} 
     \begin{minipage}  {0.135\linewidth}
        \centering
        \includegraphics [width=1\linewidth,height=1\linewidth]{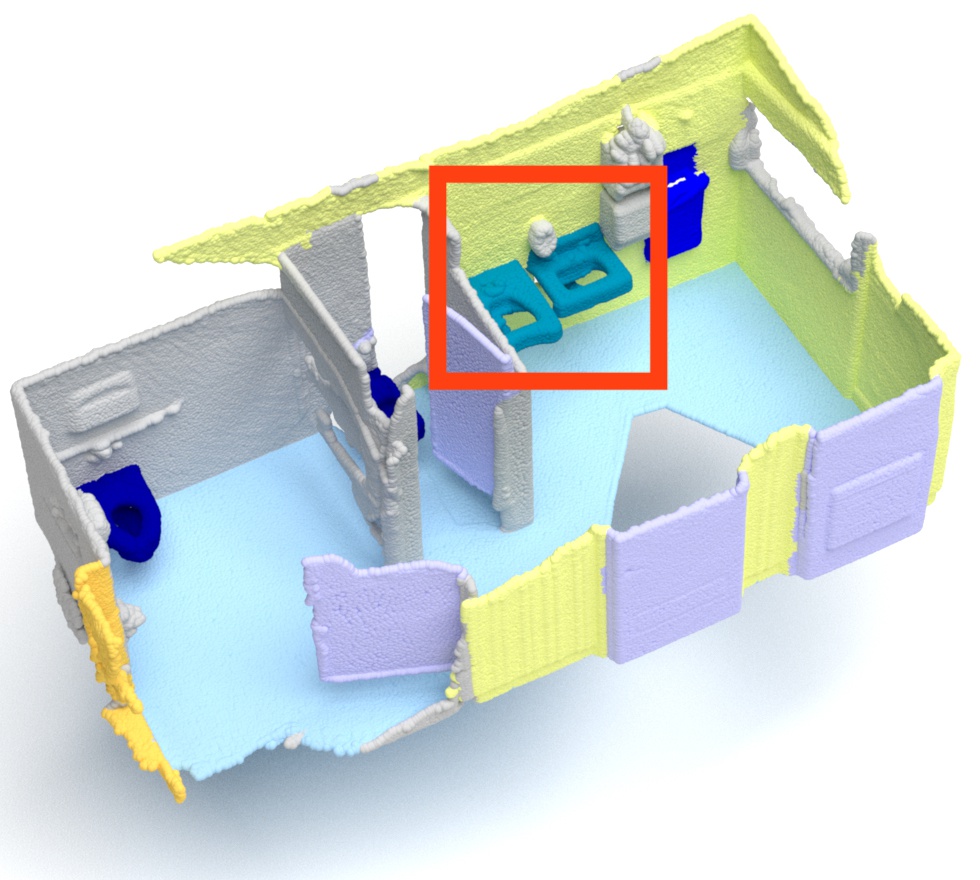}
    \end{minipage} 
     \begin{minipage}  {0.135\linewidth}
        \centering
        \includegraphics [width=1\linewidth,height=1\linewidth]{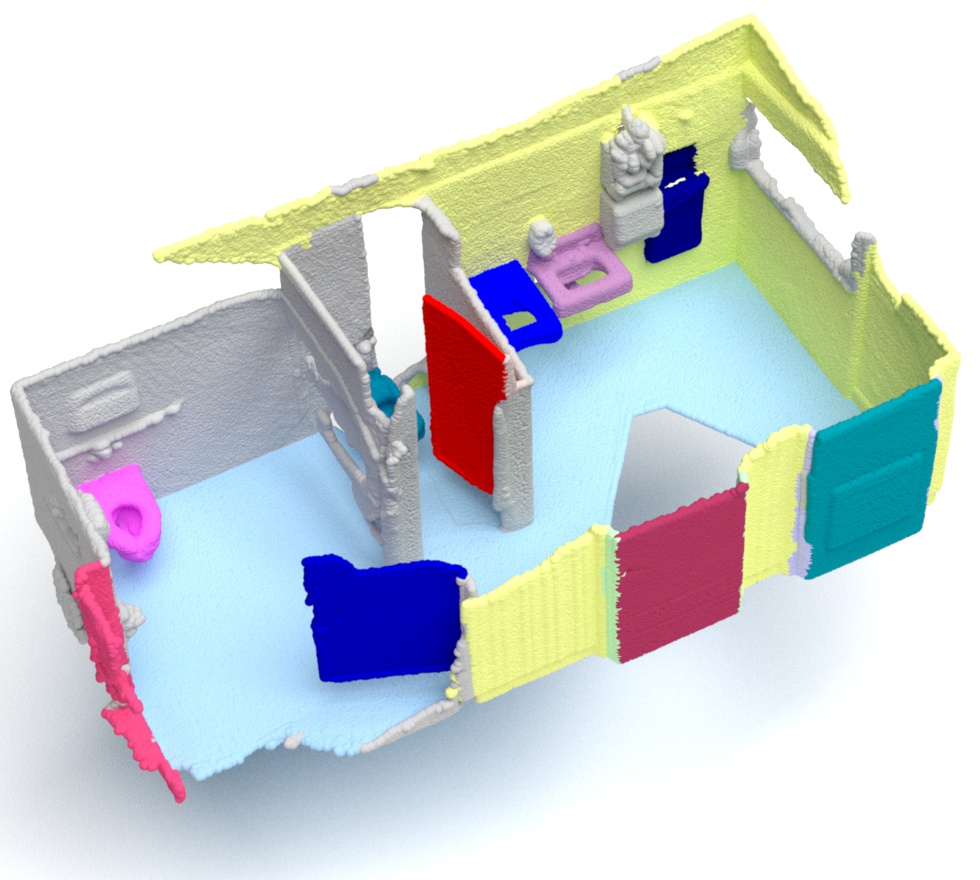}
    \end{minipage} 
  

    \begin{minipage}  {0.135\linewidth}
        \centering
        \includegraphics [width=1\linewidth,height=1\linewidth]{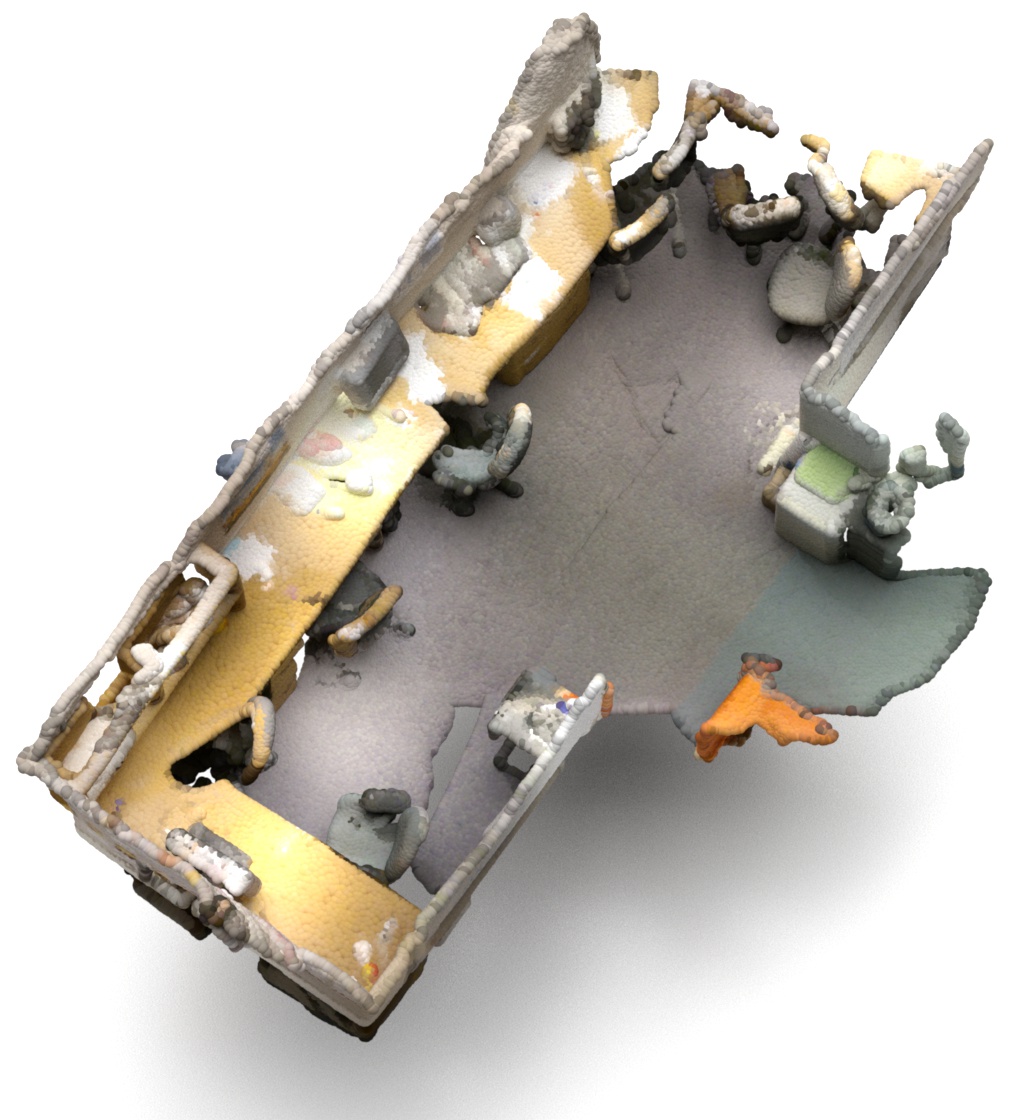}
    \end{minipage}
    \begin{minipage}  {0.135\linewidth}
        \centering
        \includegraphics [width=1\linewidth,height=1\linewidth]{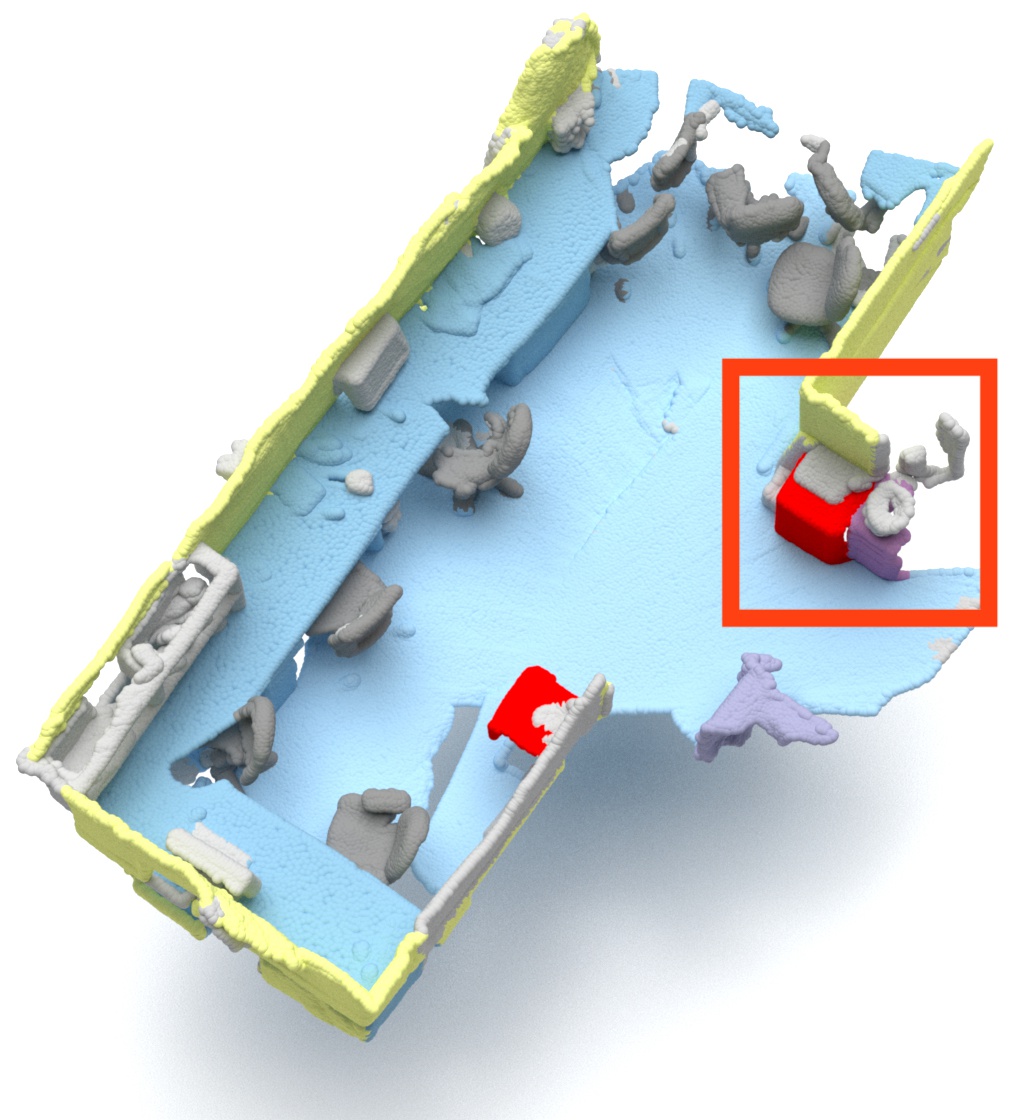}
    \end{minipage}
     \begin{minipage}  {0.135\linewidth}
        \centering
        \includegraphics [width=1\linewidth,height=1\linewidth]{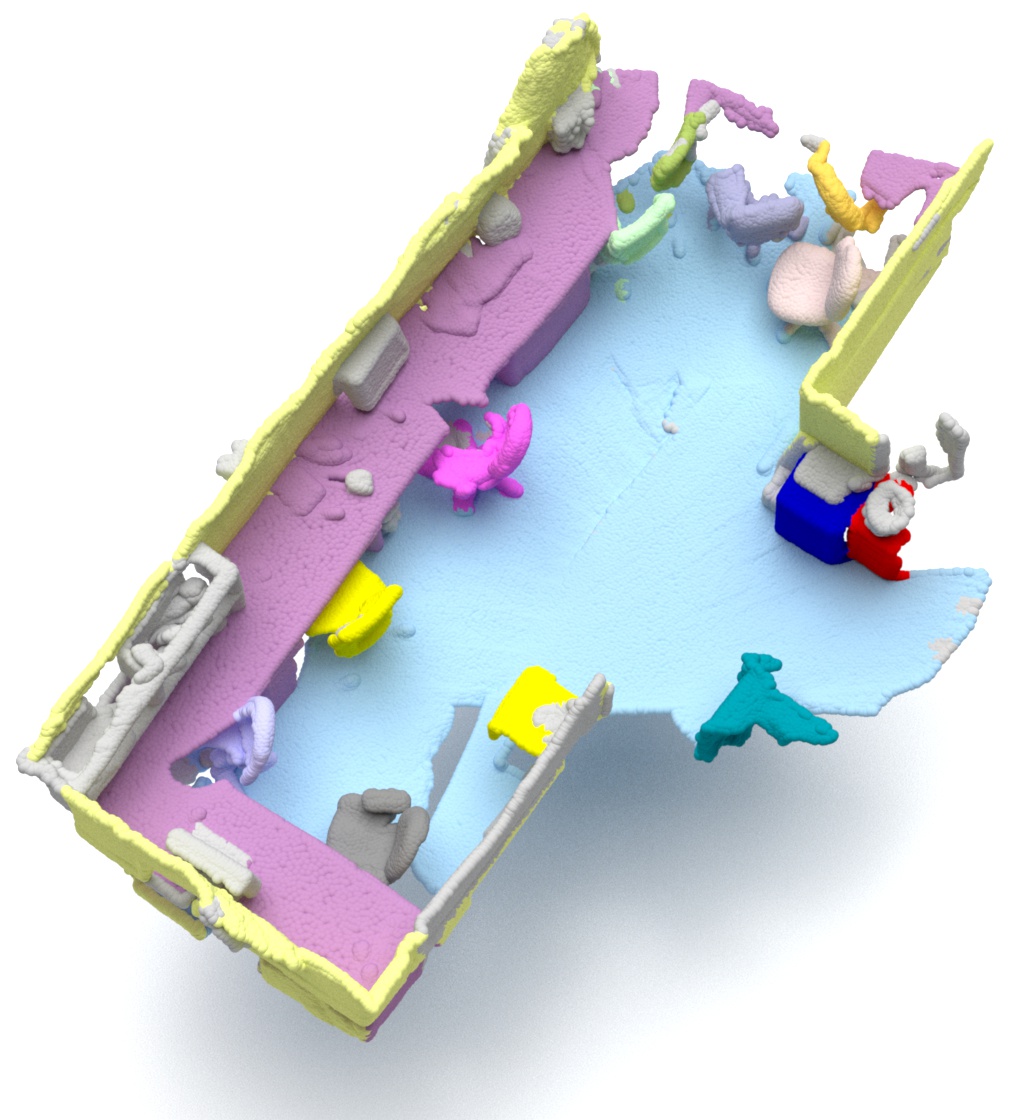}
    \end{minipage} 
    \begin{minipage}  {0.135\linewidth}
        \centering
        \includegraphics [width=1\linewidth,height=1\linewidth]{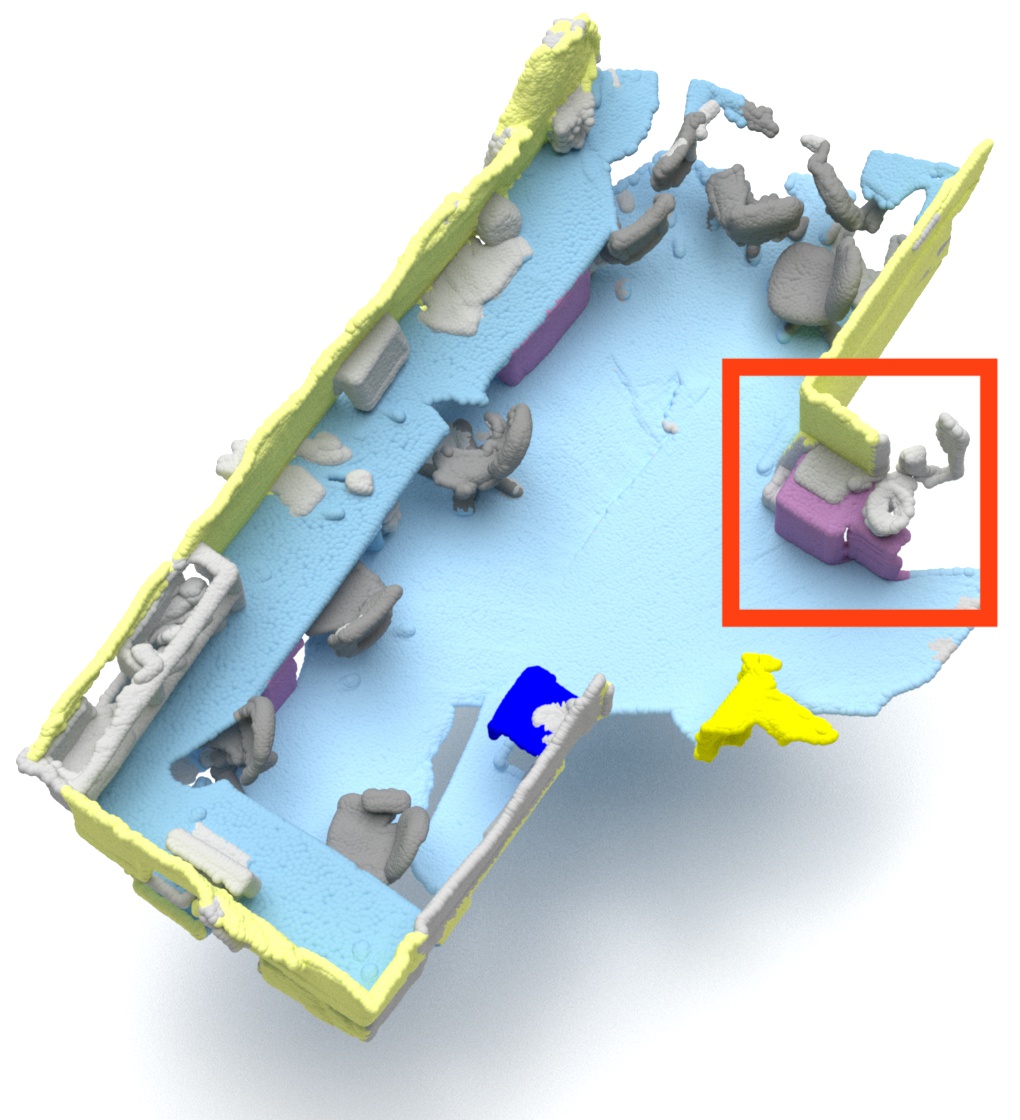}
    \end{minipage}
     \begin{minipage}  {0.135\linewidth}
        \centering
        \includegraphics [width=1\linewidth,height=1\linewidth]{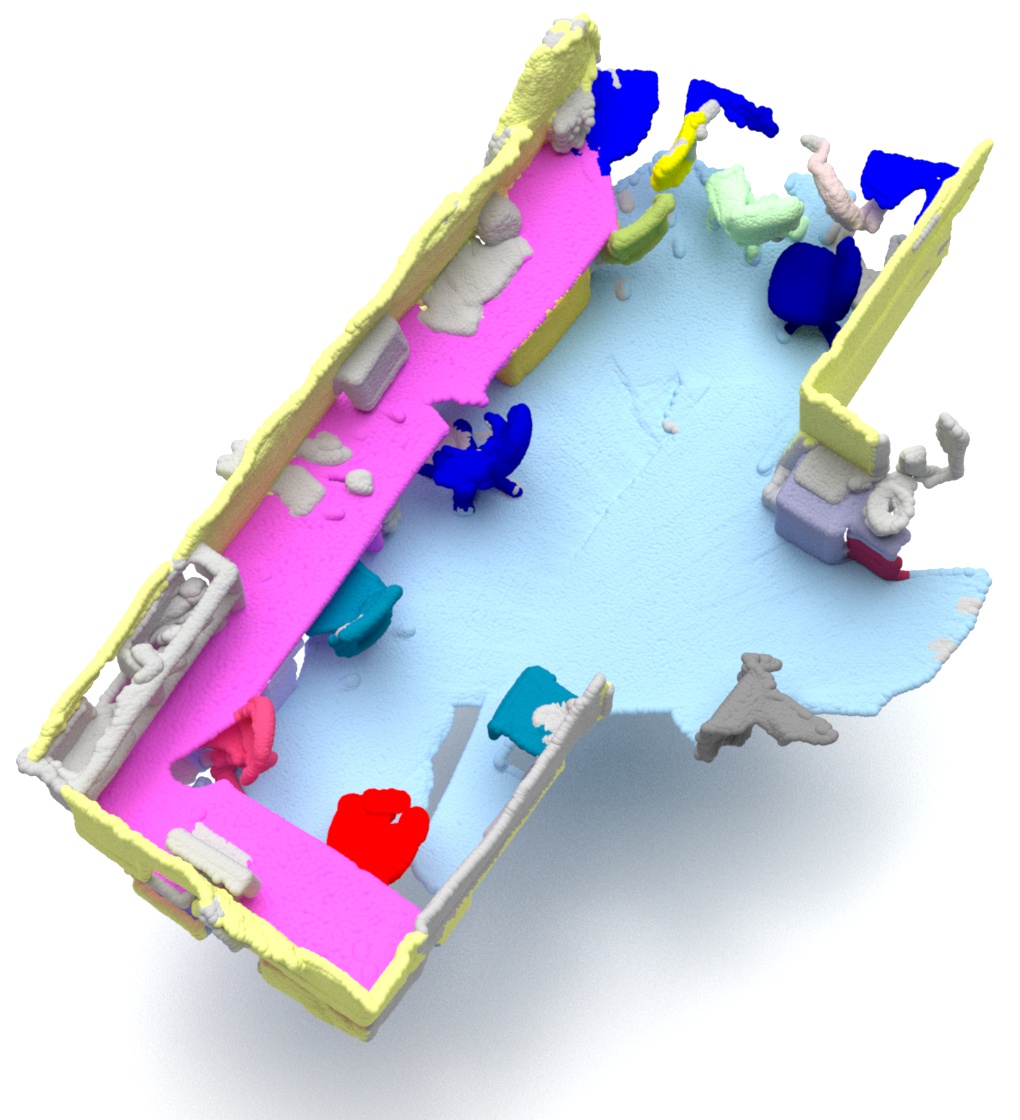}
    \end{minipage} 
     \begin{minipage}  {0.135\linewidth}
        \centering
        \includegraphics [width=1\linewidth,height=1\linewidth]{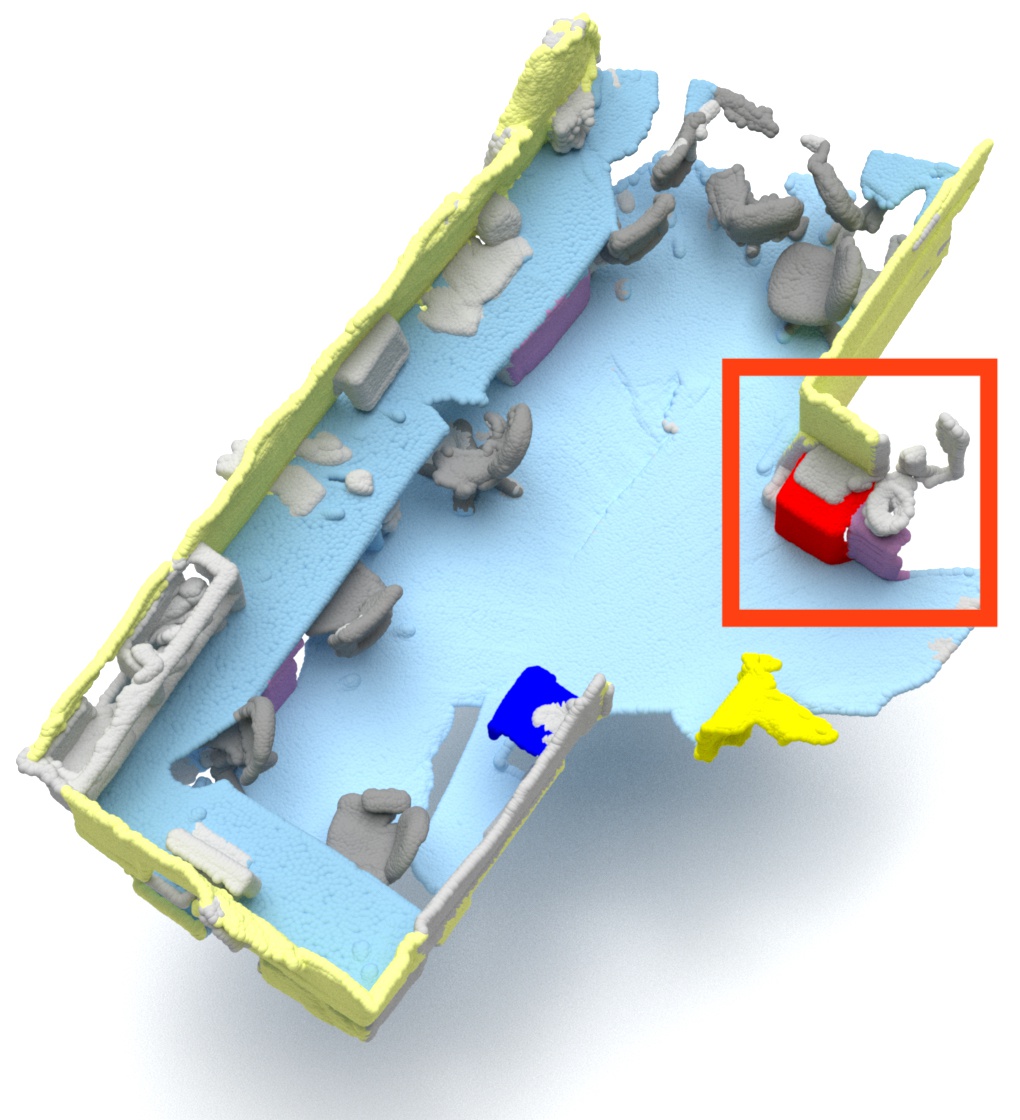}
    \end{minipage} 
     \begin{minipage}  {0.135\linewidth}
        \centering
        \includegraphics [width=1\linewidth,height=1\linewidth]{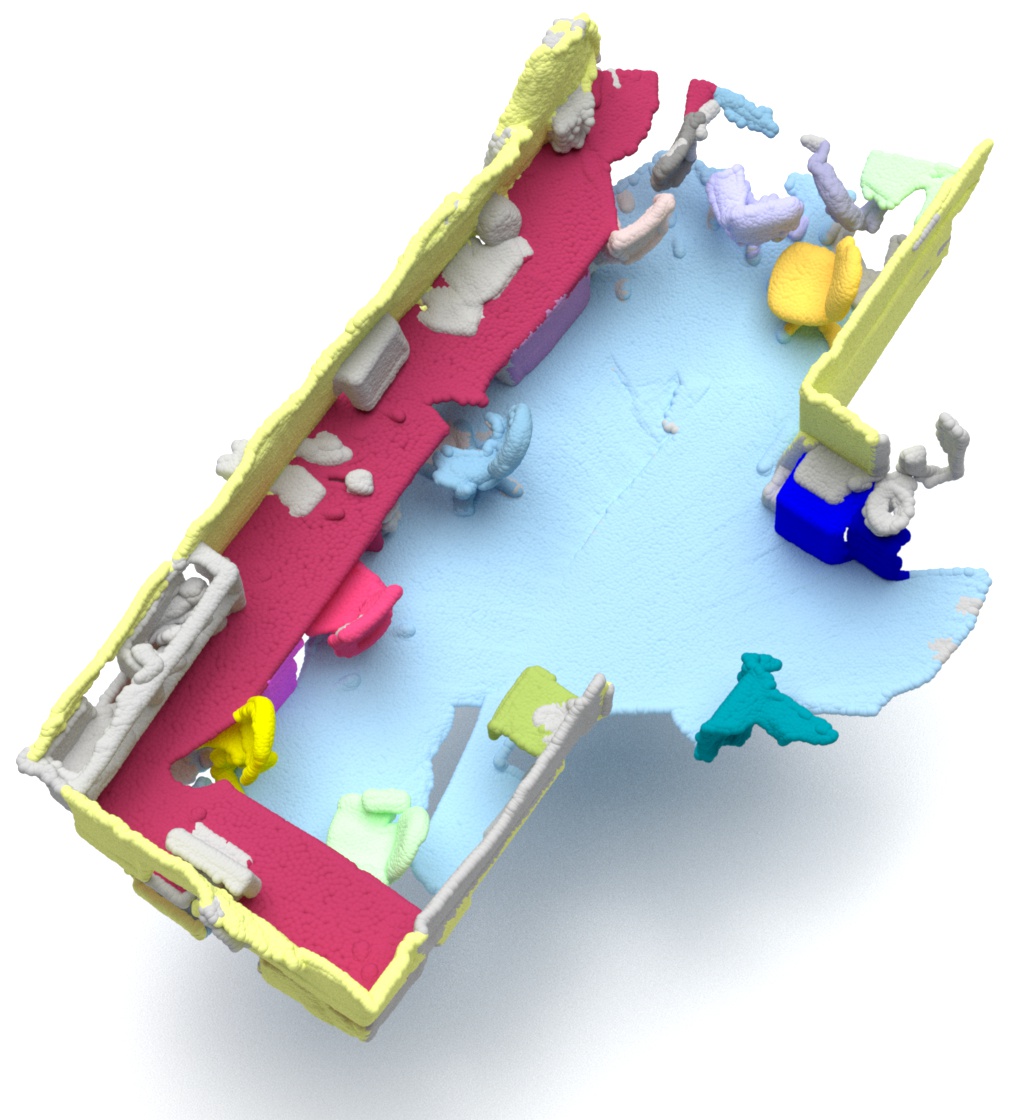}
    \end{minipage} 
  
	 
    \begin{minipage}  {0.135\linewidth}
        \centering
        \includegraphics [width=1\linewidth,height=1\linewidth]{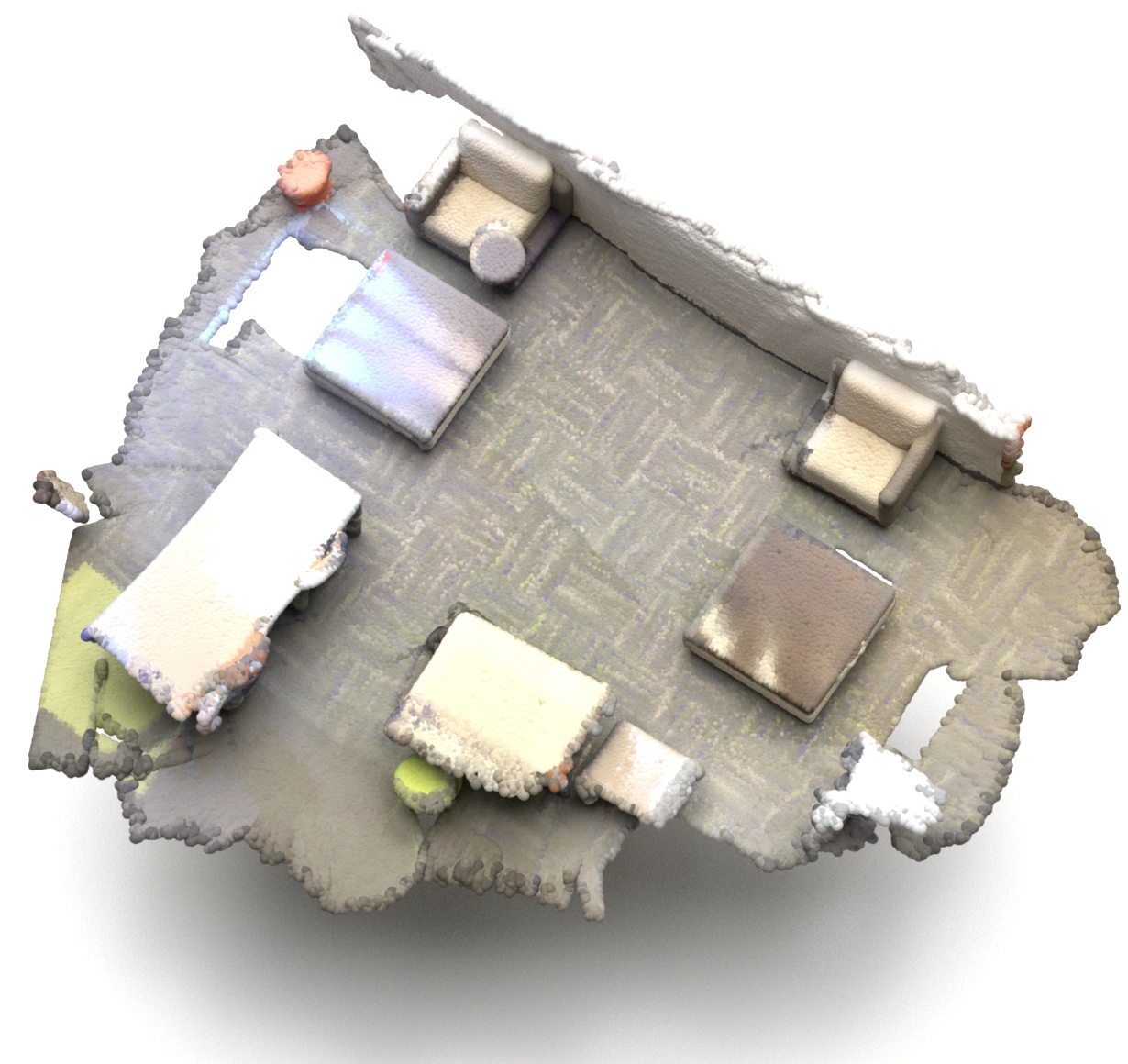}\\\footnotesize{Input}
    \end{minipage}
    \begin{minipage}  {0.135\linewidth}
        \centering
        \includegraphics [width=1\linewidth,height=1\linewidth]{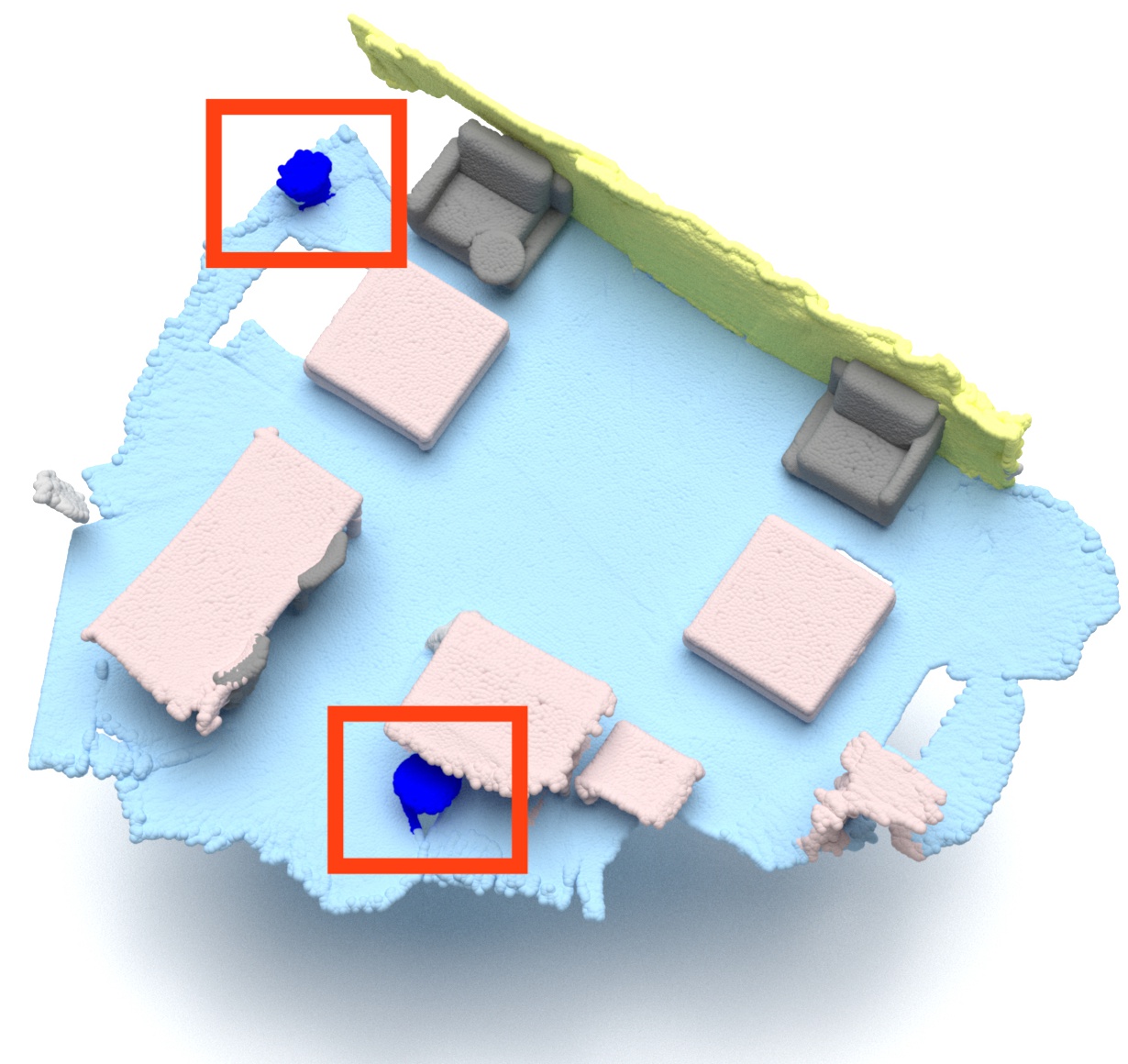}\\\footnotesize{GT Sem.}
    \end{minipage}
     \begin{minipage}  {0.135\linewidth}
        \centering
        \includegraphics [width=1\linewidth,height=1\linewidth]{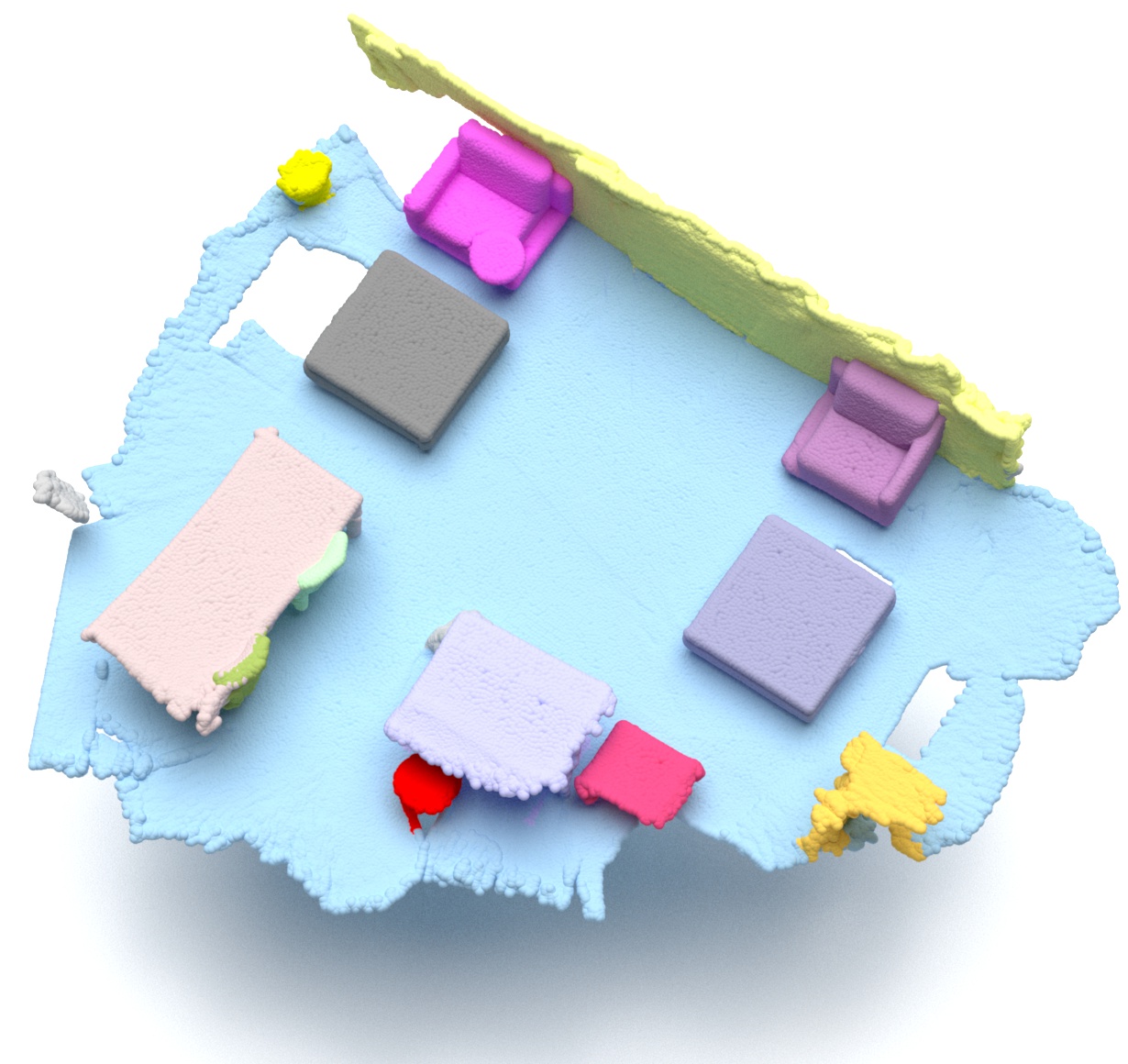}\\\footnotesize{GT Inst.}
    \end{minipage} 
    \begin{minipage}  {0.135\linewidth}
        \centering
        \includegraphics [width=1\linewidth,height=1\linewidth]{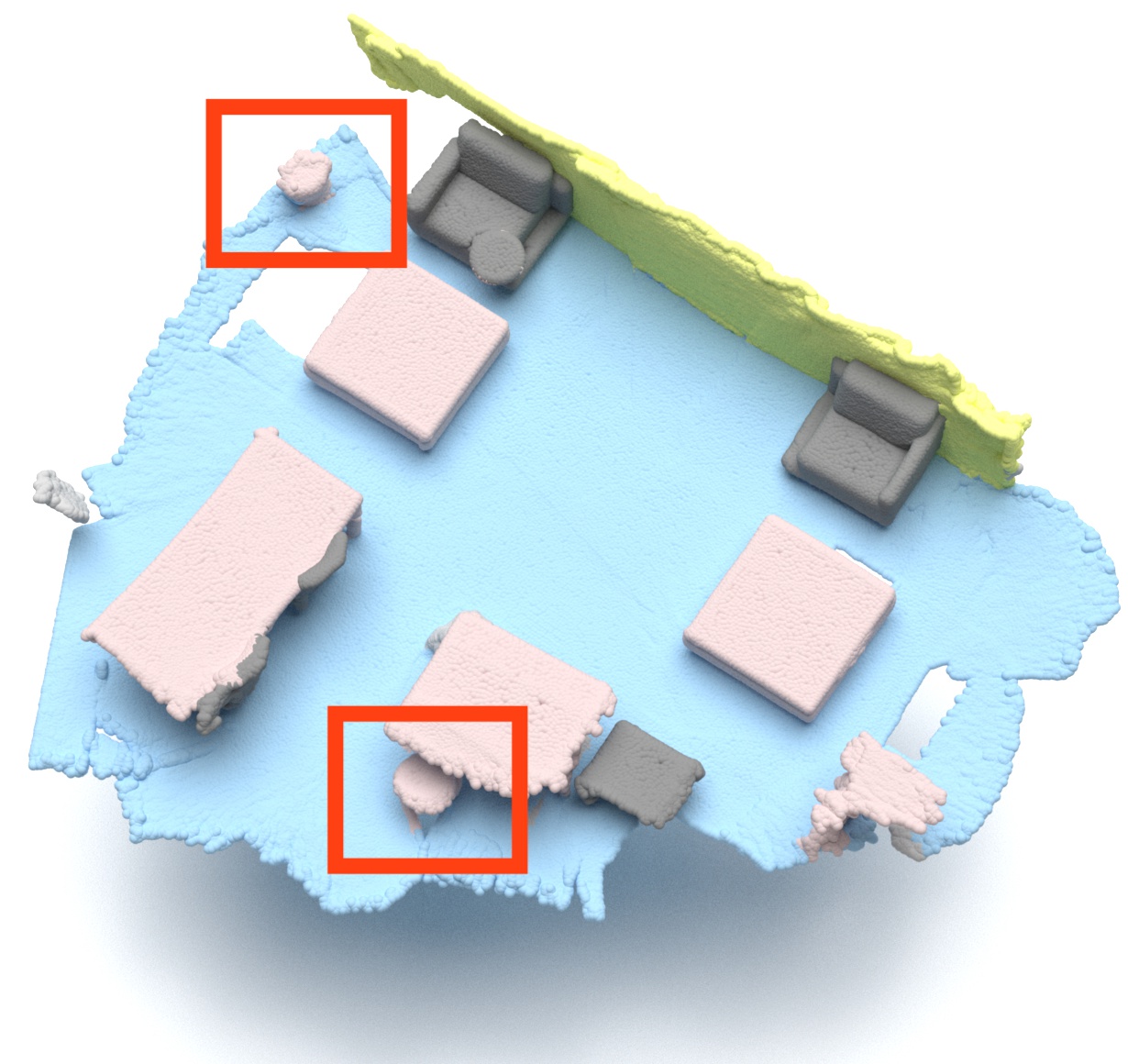}\\\footnotesize{Baseline Sem.}
    \end{minipage}
     \begin{minipage}  {0.135\linewidth}
        \centering
        \includegraphics [width=1\linewidth,height=1\linewidth]{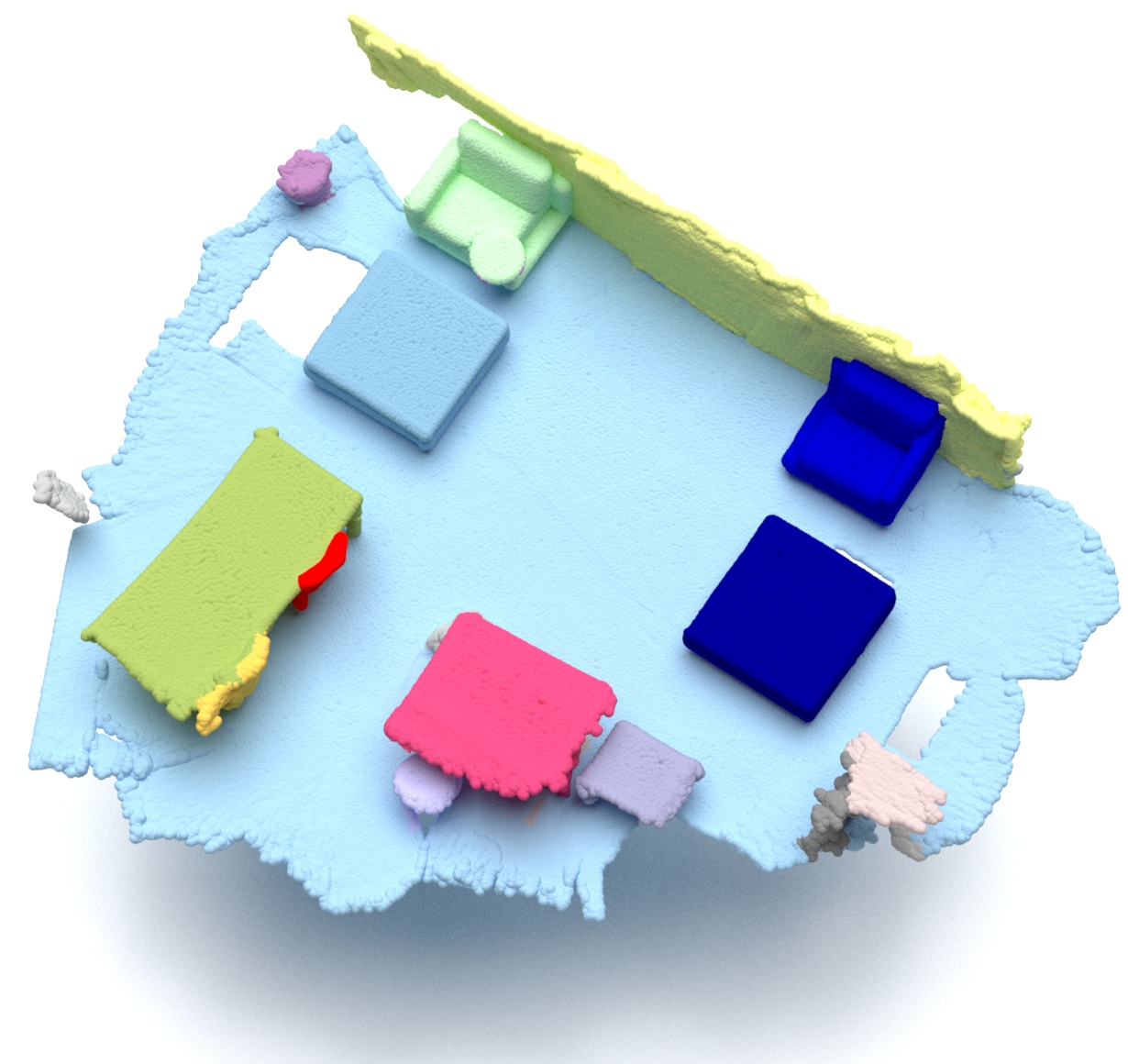}\\\footnotesize{Baseline Inst.}
    \end{minipage} 
     \begin{minipage}  {0.135\linewidth}
        \centering
        \includegraphics [width=1\linewidth,height=1\linewidth]{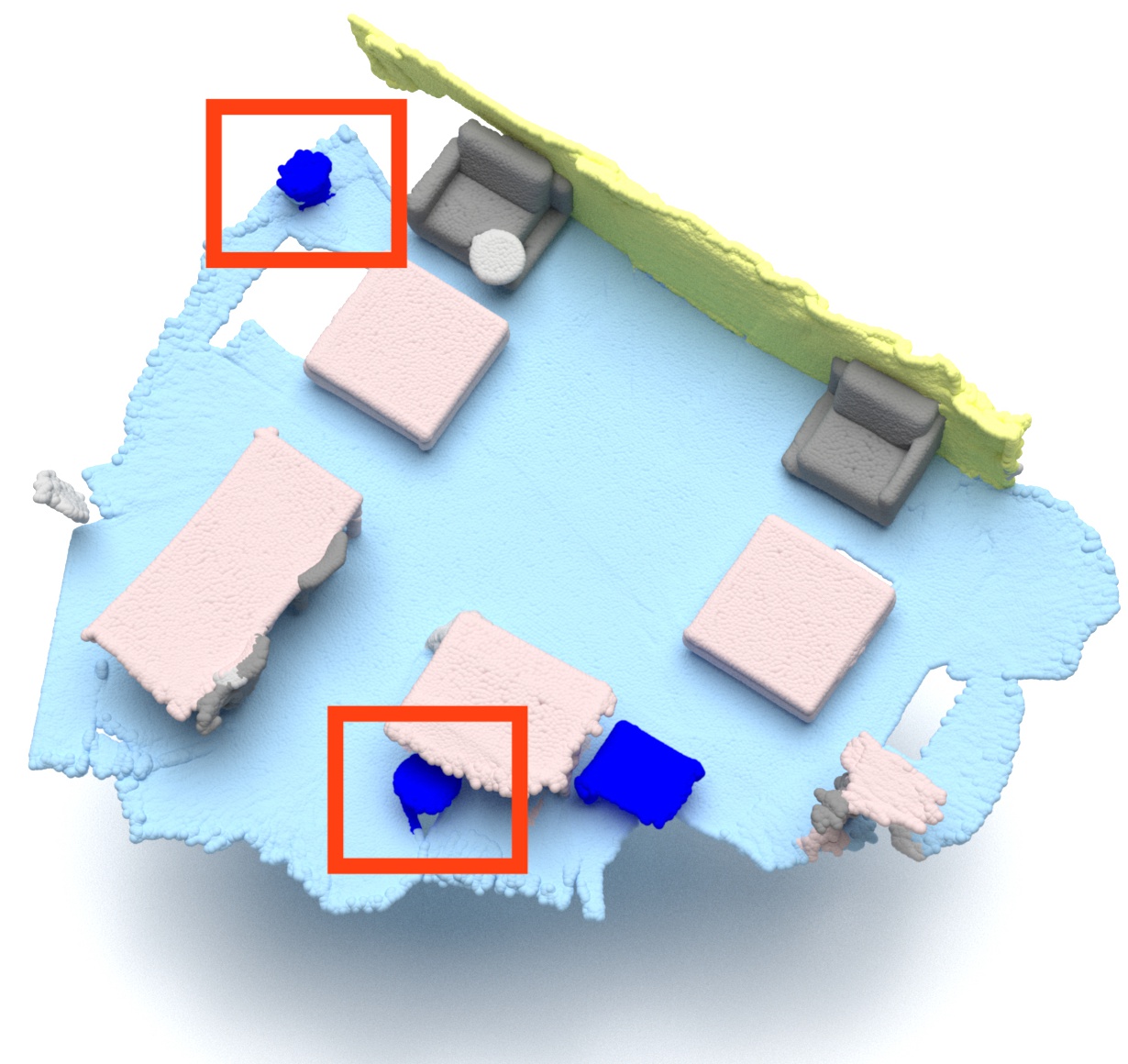}\\\footnotesize{Ours Sem.}
    \end{minipage} 
     \begin{minipage}  {0.135\linewidth}
        \centering
        \includegraphics [width=1\linewidth,height=1\linewidth]{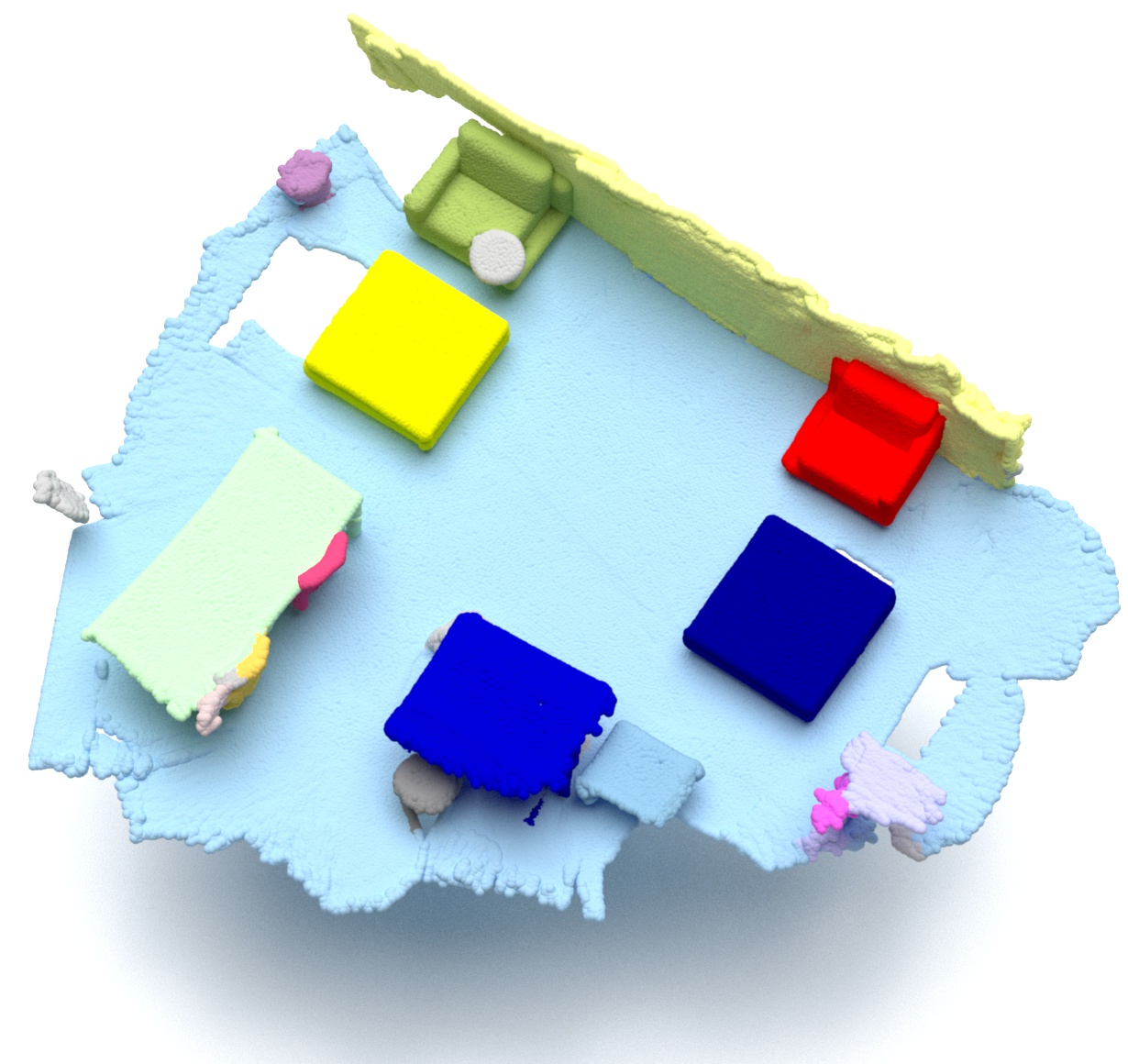}\\\footnotesize{Ours Inst.}
    \end{minipage} 
  
	 
	 \vspace{0.2cm}

    \begin{minipage}  {0.04\linewidth}
        \centering
        \includegraphics [width=0.5\linewidth,height=0.5\linewidth]
        {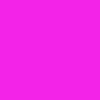}
    \end{minipage}\footnotesize bed
    \begin{minipage}  {0.04\linewidth}
        \centering
        \includegraphics [width=0.5\linewidth,height=0.5\linewidth]{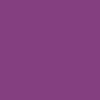}
    \end{minipage}\footnotesize cabinet
    \begin{minipage}  {0.04\linewidth}
        \centering
        \includegraphics [width=0.5\linewidth,height=0.5\linewidth]
        {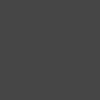}
    \end{minipage}\footnotesize chair
    \begin{minipage}  {0.04\linewidth}
        \centering
        \includegraphics [width=0.5\linewidth,height=0.5\linewidth]{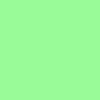}
    \end{minipage}\footnotesize counter
    \begin{minipage}  {0.04\linewidth}
        \centering
        \includegraphics [width=0.5\linewidth,height=0.5\linewidth]{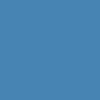}
    \end{minipage}\footnotesize desk
    \begin{minipage}  {0.04\linewidth}
        \centering
        \includegraphics [width=0.5\linewidth,height=0.5\linewidth]{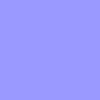}
    \end{minipage}\footnotesize door
    \begin{minipage}  {0.04\linewidth}
        \centering
        \includegraphics [width=0.5\linewidth,height=0.5\linewidth]{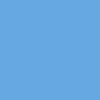}
    \end{minipage}\footnotesize floor
    \begin{minipage}  {0.04\linewidth}
        \centering
        \includegraphics [width=0.5\linewidth,height=0.5\linewidth]{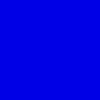}
    \end{minipage}\footnotesize otherfurniture
    \begin{minipage}  {0.04\linewidth}
        \centering
        \includegraphics [width=0.5\linewidth,height=0.5\linewidth]{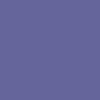}
    \end{minipage}\footnotesize sofa
    \begin{minipage}  {0.04\linewidth}
        \centering
        \includegraphics [width=0.5\linewidth,height=0.5\linewidth]{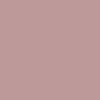}
    \end{minipage}\footnotesize table
    \begin{minipage}  {0.04\linewidth}
        \centering
        \includegraphics [width=0.5\linewidth,height=0.5\linewidth]{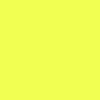}
    \end{minipage}\footnotesize wall
    \begin{minipage}  {0.04\linewidth}
        \centering
        \includegraphics [width=0.5\linewidth,height=0.5\linewidth]{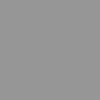}
    \end{minipage}\footnotesize ignore
    
    \caption{Visual comparison between baseline and ours (best viewed in color and by zoom-in). GT: Ground Truth. Sem.: Semantic labels. Inst.: Instance labels. The main difference is highlighted with a red bounding box. The bottom color map is for semantic labels. More examples are given in the supplementary material.}
    \label{fig:visual_comp}
\vspace{-0.4cm}
\end{figure*}

\subsection{Object Detection Results}
\label{sec:exp_det}

The instance predictions of instance segmentation can be easily transformed into bounding box predictions, by obtaining the minimum and maximum coordinates of the masked instances. We empirically find that the generated object detection results from the instance predictions work significantly better than previous methods tailored for 3D object detection in terms of mAP\textsubscript{50}, as shown in Table~\ref{tab:exp_det}. This finding also shows that our approach outputs high-quality instance segmentation results with fewer artifacts. 

\subsection{Visual Comparison}
\label{sec:visual_comp}

We visually compare our approach with previous state-of-the-art methods in Fig.~\ref{fig:visual_comp}. More examples are given in the supplementary material. The visualizations demonstrate that our method tends to correctly recognize the classes of the instances. It implies that our approach is able to generate more high-quality instance segmentation results.

\section{Conclusion}
In this work, we have presented a mask-attention-free transformer for the 3D instance segmentation task. We first observe the issue of low-recall of the initial masks in existing works. It adds training difficulty and slows down convergence. We thus avoid using mask attention and instead propose an auxiliary center regression task to guide the cross-attention. To fit center regression, we develop a series of designs. A dense distribution of position queries is learned to yield a higher recall of the perceived instances. Also, relative position encoding and iterative refinement are designed to further boost the performance. Each component is verified to be effective.

\vspace{-0.2cm}
\section*{Acknowledgements}
\vspace{-0.1cm}
This work was supported in part by the Research Grants Council under the Areas of Excellence scheme grant AoE/E-601/22-R and Shenzhen Science and Technology Program
KQTD20210811090149095.

{\small
\bibliographystyle{ieee_fullname}
\bibliography{egbib}
}

\end{document}